\documentclass[10pt,amssymb,twocolumn,notitlepage]{article}

\usepackage{epigraph}
\usepackage[affil-it]{authblk}
\usepackage{dcolumn}
\usepackage[stable]{footmisc}
\usepackage{blkarray}
\usepackage{bm}
\usepackage[mathscr]{euscript}
\usepackage[font=scriptsize]{caption}
\usepackage{subcaption,graphicx}
\usepackage[table]{xcolor}
\usepackage[margin=1.7cm]{geometry}
\usepackage{tcolorbox}
\usepackage{hyperref}
\usepackage{gensymb}
\usepackage[font=footnotesize]{caption}
\usepackage{hyperref}
\hypersetup{
    colorlinks=true,
    linkcolor=blue,
    filecolor=magenta,      
    urlcolor=blue,
}
\usepackage{steinmetz}
\usepackage{amssymb}
\usepackage{makecell}
\usepackage{algorithm,lipsum,xcolor,caption}
\usepackage{enumitem}
\usepackage{amsmath,esint}
\usepackage{fancyvrb}
\usepackage{xcolor}
\usepackage{arydshln}
\usepackage{mathtools}

\definecolor{newblue}{rgb}{0.0, 0.28, 0.67}
\definecolor{newgreen}{rgb}{0.13, 0.55, 0.13}
\definecolor{newred}{rgb}{0.87, 0.72, 0.53}

\definecolor{newblue}{rgb}{0.0, 0.28, 0.67}
\definecolor{newgreen}{rgb}{0.13, 0.55, 0.13}
\definecolor{newred}{rgb}{0.87, 0.72, 0.53}
\usepackage{stackengine} 

\newcommand{\ihat}{\boldsymbol{\hat{\imath}}}

\title{Multiset Neurons}
\author{Luciano da Fontoura Costa \\ \emph{luciano@ifsc.usp.br}}
\affil{S\~ao Carlos Institute of Physics -- DFCM/USP} 
\date{1st Nov.  2021}

\begin{document}

\twocolumn[
\begin{@twocolumnfalse}
    \maketitle
    \begin{abstract}
The present work reports a comparative performance of artificial neurons obtained in terms of the real-valued Jaccard and coincidence similarity indices and respectively derived functionals.   The interiority index and classic cross-correlation are also included  for comparison purposes. After presenting the basic concepts related to real-valued multisets and the adopted similarity metrics, including the generalization of the real-valued Jaccard and coincidence indices to higher orders, we proceed to studying the response of a single neuron, not taking into account the output non-linearity (e.g.~sigmoid), respectively to the detection of gaussian two-dimensional stimulus in presence of displacement, magnification, intensity variation, noise and interference from additional patterns.   It is shown that the real-valued Jaccard and coincidence approaches are substantially more robust and effective than the interiority index and the classic cross-correlation.  The coincidence-based neurons are shown to have the best overall performance respectively to the considered type of data and perturbations.  The potential of the multiset neurons is further illustrated with respect to the challenging problem of image segmentation, leading to impressive cost/benefit performance. The reported concepts, methods, and results, have substantial implications not only for pattern recognition and machine learning, but also regarding neurobiology and neuroscience. 
    \end{abstract}
\end{@twocolumnfalse} \bigskip
]

\section{Introduction}

A great deal of human perception and cognition, as well as of many other living beings,
critically rely on neuronal transduction and processing of several types of information.
From a simplified mathematical perspective, a neuron has been understood as a cell specialized in
processing and transmitting signals.  In a very simplified approach to modeling neuronal operation,
known as \emph{integrate-and-fire}, neuronal dynamics can be though as involving the two following main 
stages: (i) \emph{integration:} an  inner product between the input stimulus and the respective synaptic
weights, yielding an accumulated value; and (ii) \emph{fire:} the subsequent application of a non-linear function,
such as a sigmoid, over that value, eventually yielding an action potential (e.g.~\cite{haykin,mcculloch}).

This type of operation can be complemented, regarding the geometrical/shape aspects of neuronal operation,
in terms of the concept of \emph{receptive field} (e.g.~\cite{HubelWiesel,hubel}) defined with respect to some input 
stage space.  For instance, several of  the ganglion cells of the retina (e.g.~\cite{turner}) have been characterized 
by respective antagonic receptive  fields defined on the visual space (scene) or along the retina surface (retinotopic).  
Cortical neuronal cells often operate on topographical mappings of the visual field (e.g.~\cite{HubelWiesel,hubel}). 
The mathematical modeling of these receptive fields therefore provides an effective manner for representing, modeling, 
and better  understanding neuronal operation according to a systemic representation which is directly related to 
the concepts of correlation, convolution and point-spread functions (e.g.~\cite{brigham:1988,RaoHwang,oppenheim:2009,Parr:2013}).

In addition to its dynamic properties along time, the shape of receptive fields has been
understood to play an important role in detecting and processing patterns.  Indeed,
a more elaborated dendritic arborization will tend to have enhanced chances of receiving more
synaptic connections.  The importance of the neuronal geometry seems to be so important
that it often adapts to the type of function the neuron performs (e.g.~\cite{Cajal,Friedman,Grueber}).
Among the several possible interrelationships between neuronal shape and function, we have
that the alignment and similarity between the visual signal and the neuronal two-dimensional 
distribution of synaptic weights tend to result in higher neuronal activation, therefore providing some kind of
\emph{template matching} or \emph{matched filtering}.  

In the present work, we re-evaluate the functioning of single neurons in terms of
recently introducted multiset-based similarity indices capable of operating on real-valued
data~\cite{CostaSimilarity,CostaJaccard}.  More specifically, instead of using
the traditional inner product, we apply the real-valued Jaccard, interiority, and coincidence
similarity metrics~\cite{CostaSimilarity,CostaJaccard,CostaCCompl}.  

Introduced decades ago~\cite{Jaccard1}, the Jaccard similarity between two sets $A$ and
$B$ is aptly defined in terms of the following ration between set operations:
\begin{equation}
   \mathcal{J}(A,B) = \frac{|A \cap B|}{|A \cup B|}
\end{equation}

where $|A|$ stands for the cardinality of set $A$, and $0 \leq \mathcal{J}(A,B) \leq 1$.

When extended to 1D densities or non-negative functions (e.g.~\cite{jac:wiki,CostaMset,CostaJaccard,CostaSimilarity}),
the Jaccard index can be understood in particularly appealing geometrical manner as the ratio between
the area shared between the two functions and the union of their respective areas.
Of particular interest is the fact that, though extremely simple, the Jaccard index
implements an action that, though analogous to the classic inner product, is
non-linear as a consequence of the use of the maximum and minimum binary operators
which are, in multiset theory (e.g.~\cite{Hein,Knuth,Blizard,Blizard2,Thangavelu,Singh}), required for 
union and intersection of multisets, respectively.
By `binary operator' it is meant the mathematical understanding of an operation involving
two arguments.

We start by presenting the inner product, its properties,  basic multiset concepts 
(e.g.~\cite{Hein,Knuth,Blizard,Blizard2,Thangavelu,Singh}), as well as the recently 
introduced real-valued Jaccard and coincidence 
indices~\cite{CostaSimilarity,CostaMset,CostaJaccard}.  This presentation is performed first
respectively to one-dimensional input, and then extended to two- and multidimensional synaptic
inputs.    In addition to discussing the intrinsic, though limited, ability of the real product between two scalars in providing
information about their respective similarity, we also show how the 
real-valued Jaccard index can be derived in a logical manner starting from the
totally strict similarity comparison provided by the Kronecker delta function.  

Unlike in a recent study~\cite{CostaSimilarity}, which approached the subject of similarity
more generally in terms of correlation-like perspective, the neuronal perspective adopted in this work 
allowed attention to be focused on similarity comparisons where one of the arguments is kept constant,
therefore corresponding to stable synaptic weights.  In addition, for generality's sake, additional
results are reported regarding the generalization of the multiset similarity indices to higher orders, yielding
a generic similarity function that converges to the Kronecker delta product for infinite
order.

A systematic approach is then proposed and applied for comparing the performance
of neurons in pattern recognition, while adopting the standard cross-correlation as well
as the interiority, real-valued Jaccard and coincidence 
indices~\cite{CostaSimilarity,CostaMset,CostaJaccard}.  The comparison is performed
with respect to varying pattern position, intensity, scale, noise levels,  presence of
additional interfering patterns, and false positives resulting from completely  noisy data.

Several interesting results are report that, all in all, confirm that the coincidence index
provides the most strict and detailed recognition, followed by the real-valued Jaccard and
interiority indices.  The classic cross-correlation resulted almost useless for the considered
task and type of data.   These results have many implications and applications to several
related areas, some of which are also briefly discussed.

\section{The Importance of Feature Spaces}

The type of input received by a neuronal cell (biological or artificial) --- including the respective physical units,  
mutual interrelationship, and coordinate system (basis chosen for representation) ---  has a critically important 
impact on subsequent processing, including pattern recognition.  Henceforth we will understand each individual 
synaptic input to a neuronal cell as being associated to a respective \emph{measurement} or \emph{feature},
which can often be modeled as a random variable.

Figure~\ref{fig:features} depicts two main situations regarding the type of input received by a neuronal cell
(biological or artificial).  In Figure~\ref{fig:features}(a),  we have a single neuronal cell deriving its synaptic 
input directly from the coordinates of geometric data organized topographically, in which 
case there are intrinsic geometrical relationships (e.g.~continuity and proximity) implied by proximity and 
adjacency between the data elements or features that constitute potentially useful information for the neuronal 
processing.  An example of this type of input is the description of the geometry of a 3D real-world object 
(e.g.~an apple) in terms of the respective  position of each of its points, each of them corresponding to  as 
defined by an orthonormal coordinate system.  Observe that each of these features have completely homogeneous, 
having exactly the same nature (physical unit of space).  In these cases, geometric transformations such 
as translations, rotation, and eventually scaling (e.g.~\ cite{shapebook}) are well posed, and the recognition 
is often required to be performed invariant to these transformations.  

\begin{figure}[h!]  
\begin{center}
   \includegraphics[width=0.6\linewidth]{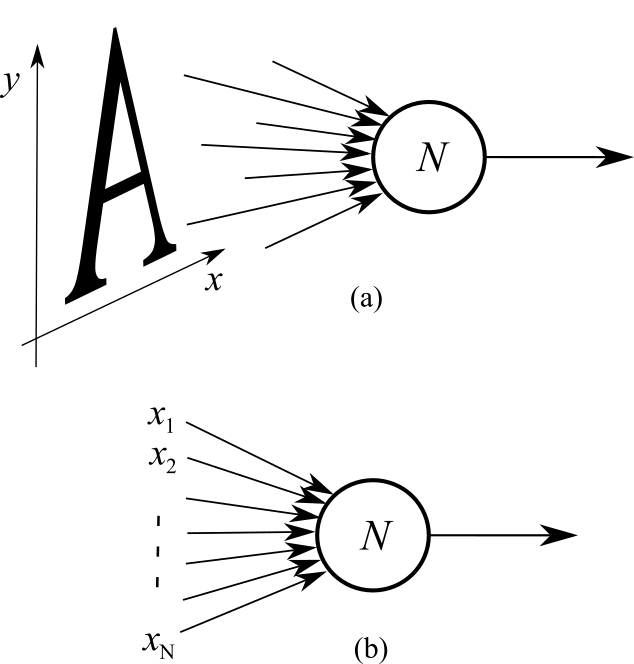}  
    \caption{The two main types of synaptic inputs to an individual neuronal cell: (a) the features
    have homogeneous units, corresponding to the positions (coordinates) of the points of the patterns to be recognized, in the
    case of this particular example being measured in
     2D  orthogonal coordinate system (the $x$ and $y$ coordinates of each of the pattern points are taken
     as respective synaptic input/feature); (b) the features are of more general and heterogeneous and
     potentially abstract and/or compound nature, such as  temperature, color, weight, age, value, etc.  Observe
     that any feature may correspond to a functional combination and/or composition of the others, while other
     features are more independent one another.  In practice, the identification of the intrinsic mathematical
     structure of the adopted features constitutes a rather challenging issue.}
    \label{fig:features}
    \end{center}
\end{figure}
\vspace{0.5cm}

However, the above discussed geometrical type of input is never verified in biological or machine recognition
systems, except for situations in which the neurons operate directly onto the visual scene projections onto the
retina, but in this case the coordinate system is no longer orthogonal or normalized (projection from 3D to 
2D).   In visual pattern recognition systems, except for the first layer receiving projections of the scene, the
neuronal input almost invariably relate to  measurements -- such as sizes, angles, areas, etc. --- that are
derived from the patterns geometry but not correspond directly to the individual position of the object points in 
an orthogonal system.

Figure~\ref{fig:features}(b) shows another situation, much more frequently observed in practice
regarding neuronal input.  Here, though the position of each feature is important, they have heterogenous
units and are not directly related coordinate systems are rather unlikely to be orthonormal, so that
invariance to rigid body transformations are unlikely to be meaningful (or viable) for neuronal response. For 
instance, the synaptic input of a specific neuronal cell may involve, respectively to the pattern being analyzed, its color, 
temperature, weight, speed, etc.  These features are rather unlikely to belong to a orthonormal coordinate system.
A better appreciation of this important effect can be obtained by considering that each feature corresponds to
a measurement that may potentially be related to the other adopted features.  These relationships can be linear,
such as when one of the features corresponds to a linear combination of other features taken or not into account,
or even non-linear combinations and compositions of features.

\begin{figure}[h!]  
\begin{center}
   \includegraphics[width=0.5\linewidth]{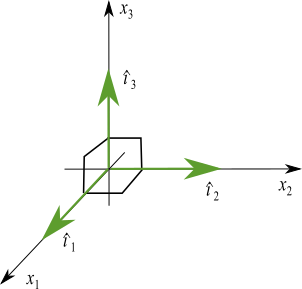}  
    \caption{The $\mathcal{R}^3$ vector space, characterized by three coordinates
    $x_1$, $x_2$ and $x_3$ associated to respective \emph{versors} $\ihat_1$, $\ihat_2$, $\ihat_3$.
    The two vectors (blue and red) are assumed to be entirely contained in the plan $(0, x_2, x_3)$.
    Each generic vector $\vec{v}$ in this space can be represented as $\left[ x_1, x_2, x_3 \right]^T =
    x_1 \ihat_1 + x_2 \ihat_2 + x_3 \ihat_3$.   }
    \label{fig:versors}
    \end{center}
\end{figure}

Figure~\ref{fig:versors} illustrates the prototypical orthonormal system in 3D, which provide an ideal reference
for synaptic input, though this is almost never observed in practical situations.  An example of real-world
situation involving this type of feature representations corresponds to the positions of objects in the 3D space,
measured by some accurate position acquisition device.  Observe that this cannot be accomplished biologically,
in the case of visual input, often rely on projections of 3D onto 2D, being feasible only by using artificial systems.

\vspace{0.5cm}

An important feature of the $\mathcal{R}^3$ space is the  \emph{orthonormality of the axes}, corresponding to 
$\left< \ihat_1,\ihat_2\right> = \left< \ihat_2,\ihat_3\right> = \left< \ihat_1,\ihat_3\right> =0$ (orthogonal) and 
$\left< \ihat_1,\ihat_1\right> = \left< \ihat_2,\ihat_2\right> = \left< \ihat_3,\ihat_3\right> =1$ (normalization).
An example of orthonormal basis for $\mathcal{R}^3$, known as \emph{canonical}, is as follows:
\begin{equation}
   \ihat_1 = \left[ 1, 0, 0 \right];  \quad \ihat_2 = \left[ 0, 1, 0 \right];  \quad \ihat_3 = \left[ 0, 0, 1 \right]
\end{equation}

In practice, patterns are characterized by a set of respective features $x_i$, $i = 1, 2, \ldots, N$ that are
pre-specified or chosen in a relatively intuitive manner, sometimes with some assistance of statistical
methods such as principal component analysis (PCA, e.g.~\cite{}).   Each of these features are typically
understood as defining each of the axis in the associated $N-$dimensional \emph{feature space}, allowing
each pattern to be mapped into a respective vector in this space. 

A critical problem not often realized, taken into account or studied, regards the fact that the so obtained
feature spaces \emph{are almost invariably non-orthonormal} as a consequence of several effects.  
  
Figure~\ref{fig:features} illustrates one such situation in which the patterns to be analyzed are
originally in an orthonormal space $\left[x_1, x_2, x_3 \right]$, where distances and geometric
transformations such as rotation are well defined and stable.  Observe that these original features
do not need to correspond to space, or even have the same physical units.  Two possible patterns are illustrated
in terms of the blue and red vectors which, for simplicity's sake, are assumed to belong to the plan
$\left[x_1=0, x_2, x_3 \right]$.   The dashed circle illustrates the position of the blue vectors that are
at distance $d$ from the red vector.  As a consequence of the orthonormality of the system $\left[x_1, x_2, x_3 \right]$,
these positions define a perfect circle, reflecting the isometry and rotational invariance of distance in an 
orthogonal coordinate system.   In the context of neuronal networks and pattern recognition, this circular
area can be immediately be related to the concept of generalization of the reference pattern corresponding
to the red vector.

\begin{figure*}[h!]  
\begin{center}
   \includegraphics[width=0.8\linewidth]{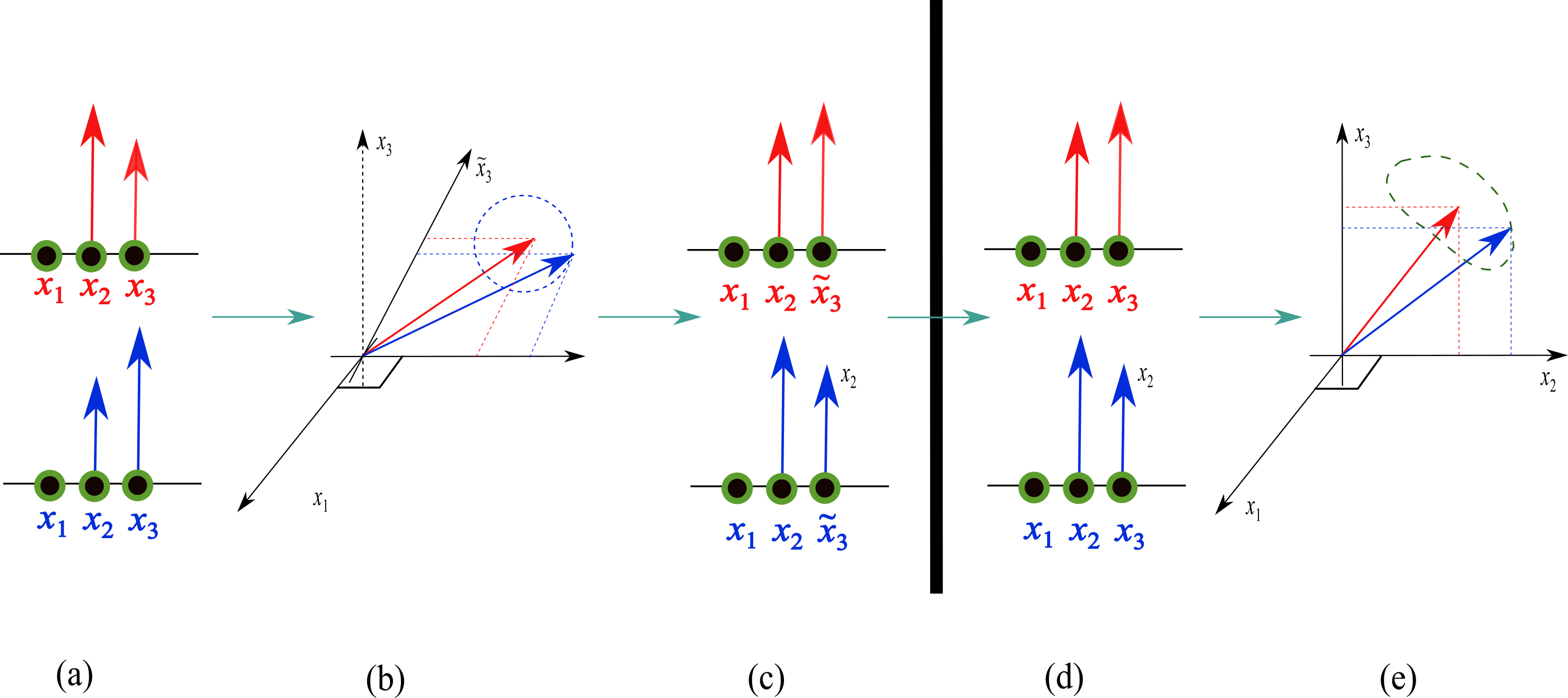}  
    \caption{The \emph{Cartesian surmise}: A real-world or abstract system (a) in which three original properties $x_1$, $x_2$ and $x_3$
    associated to an orthogonal coordinate system (b) characterize the patterns to be studied.  However, the
    measurements are  performed respectively to a different, sheared coordinate system involving 
    $x_1$, $x_2$ and $\tilde{x}_3$ (c) and then treated as if belonging to an orthogonal coordinate system (d,e).
    The horizontal bar represents the separation between the pattern and analysis domains.
    Though each of the individual measurement values will be preserved, respective interrelationships such as distances
    and magnitudes, including rotational invariance and generalization regions, can become strongly modified or invalidated. }
    \label{fig:features}
    \end{center}
\end{figure*}
\vspace{0.5cm}

However, in practice the measurements are adopted in a mostly intuitive manner while taking into account the
potential of each of them for characterizing and discriminating between the available patterns.  As such,
it is highly unlikely that the adopted features will correspond to an orthonormal coordinate system.  In the
specific case in Figure~\ref{fig:features}, one of the adopted features actually corresponds to a linear combination
of two of the original coordinates, e.g.~$\tilde{x}_3 = 2 x_3 + x_2$.   Other situations undermining the orthonormality
of the feature space include scaling of the original features or non-linear transformations of their individual or
combined values.  

In practice, the adopted features are often understood, for simplicity's sake, to be associated to an orthonormal
feature space, as illustrated in Figure~\ref{fig:features}(d,e).   This approach, which is characteristic to a large
range of approaches in pattern recognition and neuronal networks, will be henceforth referred to 
as the \emph{Cartesian surmise}.  

Though the measurement values will remain property represented
in each corresponding axis, it is no longer valid to assume rotational invariance or distance preservation respectively
to the original, real measurements  $\left[x_1, x_2, x_3 \right]$.  In Figure~\ref{fig:features}, this is illustrated by the
fact that the isometric distance circle having become a deformed region that, in this specific case, corresponds to
a shape more complex than an ellipsis, which would be otherwise obtained in case the original features had only
been scaled.   Even though generalization is still catered for, it is no longer isometric and by no means correspond
to a circle or sphere.

One of the most immediate consequences of the critically important effects of the Cartesian surmise concerns the
fact that, unless the adopted measurements do related directly to an orthonormal coordinate system, it makes
little sense to expect or implement rotational invariance in the neuronal operation because, even if this property
is observed in the adopted feature space, therefore defining isometric relationships between the vectors associated
to the patterns, this by no means translate to the original properties of the system.  Ideally, in the rather unlikely
case the interrelationship between the adopted and originally orthonormal features is known, the neurons can
have their operation designed so as to compensate for the respective divergences from orthonormality.
At the same time, it does not by any means follows that the neuronal operation can have any erratic generalization
regarding the respectively implemented quantification of similarity, in the sense that relatively symmetric generalization
regions should be sought, though not necessarily corresponding to perfect spheres.  Actually, other potentially more important
requirements can be taken into account while defining the neuronal basic operation, such as minimizing the
sensitivity of the neuronal output respective to perturbations of a single or small set of features, or normalizing
with respect to the input signal overall magnitude.

\section{Product and Similarity}

Given any two real values $x$ and $y$, their product constitutes one of the most frequently
performed algebraic operation in science and technology, not to mention daily activities.
Yet, there are some quite interesting properties of the product $xy$ that, perhaps as a consequence
of being so ubiquitous, are not commonly realized.  

Let's start with the product sign rule:
\begin{equation}  \label{eq:sign}
  \begin{array}{c c | c}  
  sign\left\{x\right\} & sign\left\{y\right\} & sign\left\{xy\right\} \\   \hline
   - & - & +  \\
   - & + & -  \\
   + & - & -  \\
   + & + & +  \\
   \end{array} \nonumber
\end{equation}

Logically, the above rules can be conceptualized as the \emph{identity} operation
of Boolean Algebra (e.g.~\cite{Hein}).

It follows that the classic product between two real values is capable of expressing
whether the two values $x$ and $y$ head toward the same direction along the real
line, in which case $sign\left\{ x y \right\} = +1$, or if they oppose one another,
yielding $sign\left\{ x y \right\} = -1$.   As such, the product operation can be understood
to quantify, in its signal, the similarity of the relative orientations of the two operands.

This important property of the classic real product hints at a yet more important
respective feature, namely the fact that \emph{the classic real product provides  
measurement of similarity} between the \emph{signs} (or direction) of two signed values 
$x$ and $y$~\cite{CostaSimilarity}. 
This particular feature of the product contributes strongly to capacity of the inner product 
for quantifying the similarity between two vectors.  More specifically, the traditional inner 
product between any two vectors $\vec{v}$ and $\vec{p}$ in an $N-$dimensional space can 
be written as:
\begin{align}
   \left< \vec{v}, \vec{p} \right> = \sum_{i=1}^N v_i p_i = \left| \vec{v} \right| \left|\vec{p} \right| \cos(\theta)
\end{align}

where $\theta$ is the smallest angle between the two vectors.  Provided the magnitudes
of $\vec{v}$ and $\vec{p}$ are kept constant, the inner product will provide an indication 
of the angular and orientation similarity between these two vectors.  Observe that the
inner product is a bilinear operation.

When translated to scalar values, the inner product becomes:
\begin{align}
   \left< x,y \right> =  x y = \left| x \right| \left| y \right| \cos(\theta)
\end{align}

which makes it clear that the scalar version of the inner product is the product of the
two scalar arguments.

In this case, the cosine similarity becomes:
\begin{align}  \label{eq:cos1D}
   \cos(\theta) = \frac{ \left< x,y \right> }{|x| |y|} = \frac{x y}{|x||y|} = \pm 1
\end{align}

Observe also that, provided $|x|\leq1$ and $|y|\leq 1$,  it will follow 
that $-1 \leq xy \leq 1$.   

In the case of $\vec{x}$ and $\vec{y}$ being vectors in $\mathcal{R}^N$, the respective
cosine similarity can be expressed as:
\begin{align}
   \cos(\theta) = \frac{ \left< \vec{x},\vec{y} \right> } { \left| \vec{x} \right| \left| \vec{y} \right| }
\end{align}

This expression implies that the cosine similarity between two vectors can be understood as corresponding to a
normalized version of the inner product between those two vectors.  As a consequence, the inner product
between two \emph{versors} (vectors with unit magnitude) is identical to the respective cosine similarity.

\begin{figure}[h!]  
\begin{center}
   \includegraphics[width=0.8\linewidth]{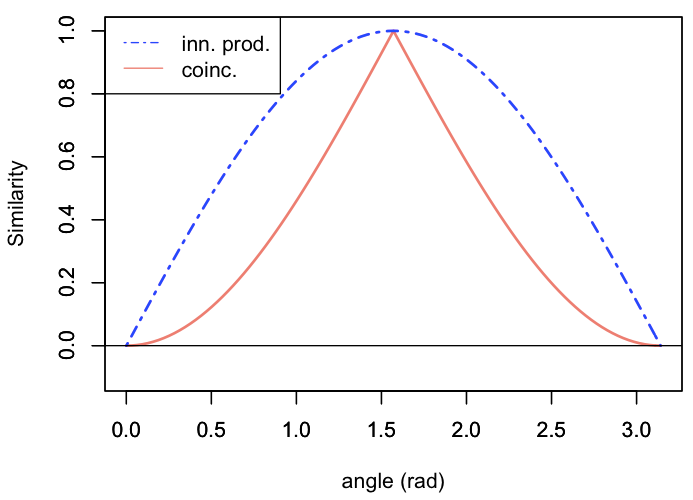}  
    \caption{The pairwise similarity between versors  in $\mathcal{R}^2$ as quantified by the
    inner product (dashed blue) and coincidence (solid salmon) similarities.  The versor $\vec{x}=[0,1]$ is compared to
    versors making angles from 0 to $\pi$ with the horizontal axis.  As expected, both similarity indices reach their
    respective peak at $\pi/2$, but the inner product similarity, which in the case of versors is identical to the cosine
    similarity, is markedly less strict in comparing similarity, providing substantially higher values than the coincidence similarity
    in all cases except for angles $0$ and $\pi/2$.}
    \label{fig:angle_sims}
    \end{center}
\end{figure}

Despite its intrinsic ability for quantifying similarity between the sign of values, 
as well as it extensive application in operations as the inner product, the real product has two important
shortcomings.  First, it is relatively difficult to be implemented in computational hardware
or even in analog circuits.  Second, it has been shown that the real product tends to be
too tolerant regarding the provided indication of 
similarity~\cite{CostaJaccard,CostaSimilarity,CostaComparing}, as illustrated in Figure~\ref{fig:angle_sims}
with respect to comparison between versors.

\vspace{0.5cm}

The substantially high similarity values generated by the inner product similarity between two vectors immediately implies that 
that it tends to provide a relatively coarse, or little strict, quantification of the relationship between the two vectors.
Another feature of particular interest in the similarity profiles such as those in Figure~\ref{fig:angle_sims} concerns
the quality factor (in analogy with filter theory, e.g.~\cite{kumar:2009}) of the quantification, which can be understood as being 
proportional to the peak value divided by the standard deviation or other dispersion measurements, therefore providing
an indication of the sharpness of the profile.  Another important property of a similarity profile is its magnitude of the derivative at and
around its peak.  Regarding the former, we have that it is zero for the inner product similarity (as expected with any smooth
function) and infinite for the coincidence.  At the same time, the magnitude of the derivatives around the peak are very small in the
case of the inner product similarity, and particularly high for the coincidence similarity.  The fact that the derivative of the
coincidence similarity profile diverge at its peak can be easily circumvented, if necessary for analytical and theoretical
studies, by representing the profile in terms of a truncated Fourier series, which is necessarily analytical (has any derivative).

One problem of having a smooth (or `blunt') similarity peak, as is the case with the inner product similarity, consists in the fact that the 
identification of its position by using derivative is highly susceptible to any level of noise.  That is so because the derivatives 
at and around the peak have very small magnitudes (smooth) and can therefore be severely disturbed by the noise during the
derivative, as this operation emphasized the high frequency content of the curve.  The sharp and intense derivative peak
resulting from the coincidence can hardly have its correct position disturbed by any reasonable level of signal noise.

On the other hand, a smoother similarity comparison profile tends to favor \emph{generalization} of the comparison, a property
that is often expected at some level in neuronal networks and pattern recognition.  In the case of Figure~\ref{fig:angle_sims}, the
relatively higher similarity values provided by the inner product, respectively to the coincidence, similarity means that input
patterns that are more different to the one used as a reference (or template) will imply larger similarity values, therefore implying
larger generalization.  Observe that the generalization property is opposite to accuracy in the similarity comparison, which means
that either one of these two properties is prioritized, or a suitable balance between them needs to be achieved.  To any extent,
as it can be appreciated from Figure~\ref{fig:angle_sims}, the coincidence similarity already presents a substantial ability for
generalization, yielding substantially high (though much smaller than the cosine similarity) for input with angles reasonably near
$\pi/2$.

Table~\ref{tab:comp} provides a qualitative relative comparison between the several properties respectively characterizing
the inner product (or cosine) and coincidence similarities.  

\begin{table}[h!]
\centering
\renewcommand{\arraystretch}{1}
\begin{tabular}{|| c | | c  | c ||}  
 \hline
 \emph{property} & inner product & coincidence \\   \hline \hline
 \emph{values} & typ.~higher & typ.~lower \\  \hline
 \emph{peak shape}  & smooth  & sharp \\ \hline
 \emph{peak concavity}  &  convex &  concave \\ \hline
 \emph{strictness} & lower & higher \\ \hline
 \emph{peak localiz.~accur.} & lower & higher  \\  \hline
 \emph{magn.~deriv.~at peak} & lower  & higher \\ \hline
 \emph{deriv.~at peak} & 0 & $\infty$ \\ \hline
 \emph{quality factor}  &  lower & higher  \\  \hline
 \emph{false neg.~prob.} & lower & higher \\ \hline
 \emph{false pos.~prob.} & higher & lower \\ \hline
 \emph{generalization} & higher & lower \\   \hline
 \emph{complexity} & higher & lower \\ \hline
\end{tabular}
\renewcommand{\arraystretch}{1}
\caption{Relative comparison of the main properties of the inner product (cosine) and coincidence similarities.
The abbreviation  \emph{peak localiz.~accur.} means peak localization accuracy, and the
term \emph{complexity} property related to conceptual and implementation (software or hardware) aspects.}
   \label{tab:comp}
\end{table}

However, being more or less strict does not necessarily imply an advantage or shortcoming of a given similarity index,
unless these trends are extreme.  Indeed, there are situations in which it may be interesting to implement a more yielding
quantification, which tends to favor false negatives.  At the same time, a more  strict similarity quantification can be particularly 
interesting in other situations in which false positives have higher costs and need to be minimized and/or enhance accuracy 
is expected in the quantification.   In summary, the localization of the position of the peak tends to be substantially more
robust and accurate in the case of the coincidence than the inner product similarity.

For these reasons, it becomes particularly interesting to
consider similarity measurements involving one or more parameters that can be used to control how strict the
quantification is, so that this can be adapted to a wide range of situations and applications.   Approaches such as
\emph{multiresolution} or \emph{multiscale} have been developed with that finality in mind.  The present work
describes a related approach in which a parameter $D$ is used to control how strict the Jaccard and coincidence
similarity indices are.

\section{Multiset Similarities in One-Dimensional Spaces}

\begin{figure}[h!]  
\begin{center}
   \includegraphics[width=0.6\linewidth]{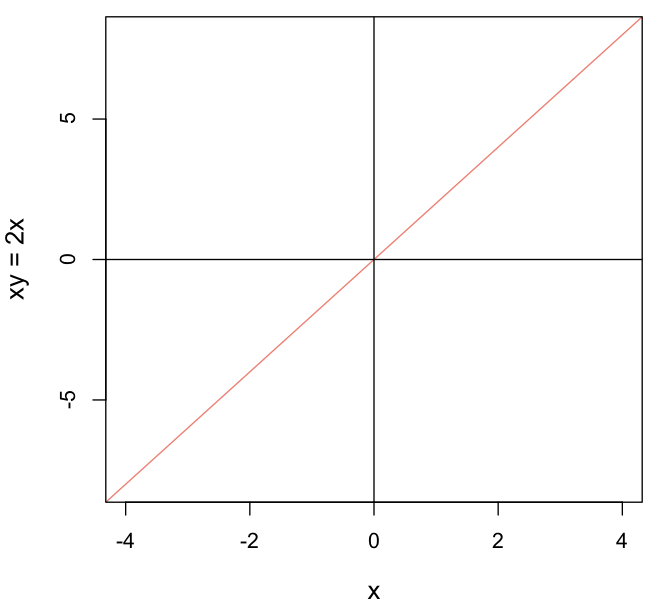}  
    \caption{The quantification of similarity between two real values as
    implemented by the real product $xy$, with $y=2$ has severe limitations. The
    curve should indicate how similar to $2$ the values of $x$ are, but monotonically
    increasing values are obtained instead.}
    \label{fig:cl_prod}
    \end{center}
\end{figure}

The considered neuronal application of similarity comparison addressed in
this work provides an interesting perspective from which to address this issue
and its related aspects.  More specifically, if we consider that the similarity is 
to be measured between the synaptic weight $y$ and respective input $x$, we 
can simplify the otherwise binary operation as an operation only on $x$ 
(i.e.~a function of $x$), with $y$ being understood as a parameter.  
Figure~\ref{fig:cl_prod} illustrates the real product seen from this perspective,
assuming synaptic weight $y=2$.

\vspace{0.5cm}

This result well illustrates the limitation of the traditional real product for quantifying
similarity.  Though the similarity will increase for $|x|$ increasing from $0$ to $1$,
it will continue to increase thereafter.  In fact, we have that:
\begin{equation}
  \lim_{x \rightarrow \infty} xy = \infty
\end{equation}

with $y=2$.

As developed in~\cite{CostaSimilarity}, the prototypical function for quantifying similarity
in the most strict manner possible consists in the Kronecker delta function, which can
be written as:
\begin{align}
   \delta_{x,y} = 
   \left\{  \begin{array}{l}
   1  \quad \emph{whenever } x=y \\
   0 \quad \emph{otherwise}
   \end{array} \right.
\end{align}

Though this function cannot provide information about the alignment of the values
$x$ and $y$, it can be readily generalized as:
\begin{align}
   \tilde{\delta}_{x,y} = 
   \left\{  \begin{array}{l}
   1  \quad \emph{whenever } x=y \\
   -1 \quad \emph{whenever } x = -y \\
   0 \quad \emph{whenever } |x| \neq |y| \\
   \end{array} \right.
\end{align}

\begin{figure}[h!]  
\begin{center}
   \includegraphics[width=0.6\linewidth]{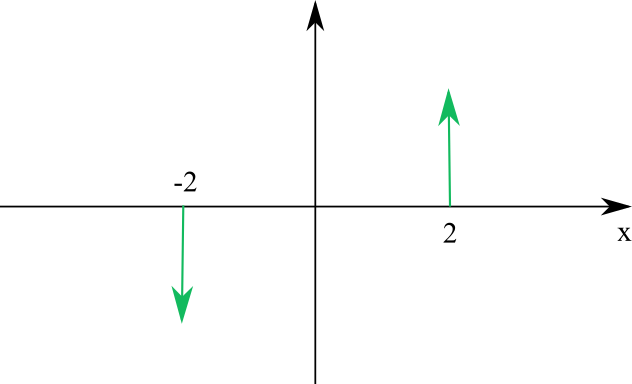}  
    \caption{The Kronecker function modified to take into 
    account the relative alignment of the real values $x$ and $y=2$,
    which is now reflected in the respective sign.}
    \label{fig:Kronecker}
    \end{center}
\end{figure}

Figure~\ref{fig:Kronecker} illustrates this function for $y=2$, i.e.~$\tilde{\delta}_{x,2}$.

\vspace{0.5cm}

\begin{figure}[h!]  
\begin{center}
   \includegraphics[width=0.6\linewidth]{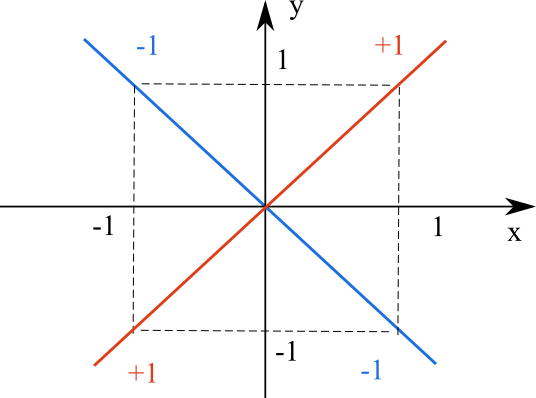}  
    \caption{The binary Kronecker function modified to take into 
    account the relative alignment of the real values $x$ and $y$.}
    \label{fig:Kron2}
    \end{center}
\end{figure}

The binary Kronecker function generalized to quantify signed similarity is shown in Figure~\ref{fig:Kron2}.

\vspace{0.5cm}

The problem with this generalized Kronecker delta function is that it is simply too
strict in its evaluation of the similarity between two real values $x$ and $y$.

We have from Equation~\ref{eq:cos1D} that, when applied on real values $x$ and $y$ in $\mathcal{R}$, 
the cosine similarity effectively acts as the Kronecker delta function on those two values, therefore
presenting rather limited potential for comparing the similarity between $x$ and $y$.

Interestingly, there is another particularly interesting possibility to quantify the similarity
between two real values taking possibly negative values~\cite{mirkin,Akbas1,Akbas2,CostaSimilarity}.
In particular, the basic scalar version of the operator in~\cite{Akbas1,Akbas2} follows 
the same sign rules as the above discussed inner product while involving only the minimum 
binary operation:
\begin{equation}
  x \sqcap y =  s_{xy} \min\left\{ s_x x, s_y y\right\}
\end{equation}

where $s_x = sign(x)$, $s_y =sign(y)$, and $s_{xy}$ is the \emph{conjoint sign function}
$s_{xy} = s_x s_y$. 

This operator has also been verified~\cite{CostaGenMops,CostaMset,CostaSimilarity} to correspond to the signed
intersection between multisets taking real, possibly negative values, being dierctoy related to real-valued
adaptations of the Jaccard similarity index. In particular,
it can be understood as a modification of the intersection between two functions in order
to consider the common or shared area of the functions with respect to the horizontal axis.

Interestingly, this product has surprising properties, including: (i) it is extremely simple
to be implemented~\cite{Akbas1,Akbas2}, e.g.~in analog electronic circuits~\cite{CostaElectronic}; 
(ii) it is conceptually simple; (iii) it obeys the sign rules in Equation~\ref{eq:sign}; (iv) its magnitude
is bound by the absolute value of the minimum between $x$ and $y$; (v) unlike the cosine similarity
when applied to 1D spaces ($x, y \in \mathcal{R}$), the real-valued Jaccard index is not limited to
yielding $\pm1$ values, but is instead capable of providing a detailed quantification of the respective
similarity.

\begin{figure}[h!]  
\begin{center}
   \includegraphics[width=0.6\linewidth]{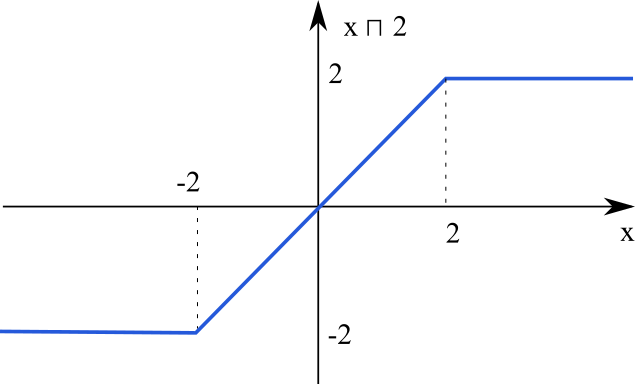}  
    \caption{The operation $x \sqcap y$ assuming $y=2$, i.e.~$x \sqcap 2$.
    Now we have that $-2 \leq x \sqcap 2 \leq 2$  The obtained profile can
    be understood as being analogous to the `receptive field' of a respective
    neuron operating in a one-dimensional input space.}
    \label{fig:common}
    \end{center}
\end{figure}

It is therefore interesting to consider this function from the neuronal perspective, i.e.~with
one of its values kept constant so as to correspond with the respective weights.  
Figure~\ref{fig:common} illustrates the operation $x \sqcap y$ for $y=2$, which can 
be understood as being analogous to the `receptive field' of a respective neuron
in the one-dimensional space $\mathcal{R}$.

\vspace{0.5cm}

It is now clear that this operation, when one of its argument is kept constant,
corresponds to a clipped version of the real product $2x$ as in our previous
example.  The saturation observed for $x>2$ is a critical feature in which it implies $x \sqcap y$ 
to become bound by the fixed value.

However, maximum similarity will be observed for any value of $x$ larger than $2$.
An interesting manner to circumvent this saturation problem consists of adopting the
following normalization:
\begin{equation}  \label{eq:Jaccard}
  J(x, y) = \frac{ s_{xy} \min\left\{ s_x x, s_y y\right\} } {\max\left\{ s_x x, s_y y \right\}} =
  \frac{f \sqcap g}{f \tilde{\sqcup} g}
\end{equation}

so that $-1 \leq x \sqcap y \leq 1$.    This normalized version of the operation $x \sqcap y$ has
been verified to correspond to the \emph{real-valued Jaccard index} applied to two real
scalar values~\cite{CostaSimilarity,CostaMset,CostaJaccard}.

Observe that, as with the standard Jaccard similarity index, the respective multiset version above
capable of operation on real values is not defined for the comparison between two null sets or
feature vectors, as it diverges to $0/0$.

\begin{figure}[h!]  
\begin{center}
   \includegraphics[width=1\linewidth]{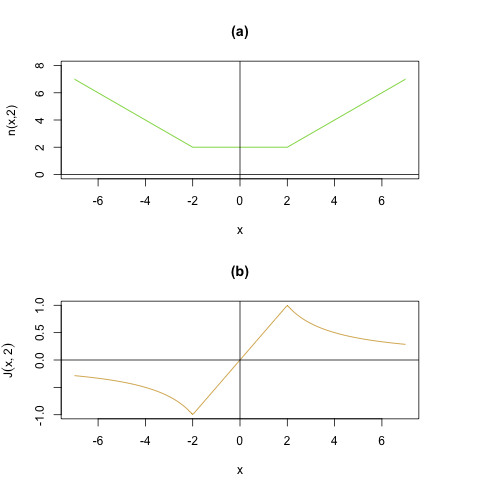}  
    \caption{The normalizing function $n(x,2) = \max\left\{ s_x x, 2 \right\}$
    and the normalized operator which corresponds to the real-valued Jaccard
    index $J(x,2)$ applied to real scalar values.}
    \label{fig:jacc}
    \end{center}
\end{figure}

Figure~\ref{fig:jacc} illustrates both the function $n(x,y) = \max\left\{ s_x x, s_y y \right\}$
and the resulting real-valued Jaccard index.

\vspace{0.5cm}

The normalizing function that constitutes the denominator has a direct correspondence 
with the generalized multiset concept of absolute union~\cite{CostaGenMops}.  Observe 
that this function increases linearly with $x$.  As a consequence, the division by the
normalizing function will penalize the similarities for $|x|>2$, yielding to two respective
peaks in the real-valued Jaccard index $J_R(x,2)$, providing enhanced quantification
selectivity.  Interestingly, this resulting index
can therefore be understood as a less strict version of the generalized Kronecker
delta (compare Figs.~\ref{fig:Kronecker} and~\ref{fig:jacc}b), while being also
more strict than the real product (as a consequence of the saturation).

The developments presented above make it clear that it is possible to define an infinity
of other similarity indices.  For instance, it is possible to control the sharpness of the
similarity peaks by using other products and normalizing functions.  

As an example, if even sharper peaks are required, we can make:
\begin{equation}
  J_3(x, y) = J(x,y)^3
\end{equation}

\begin{figure}[h!]  
\begin{center}
   \includegraphics[width=0.8\linewidth]{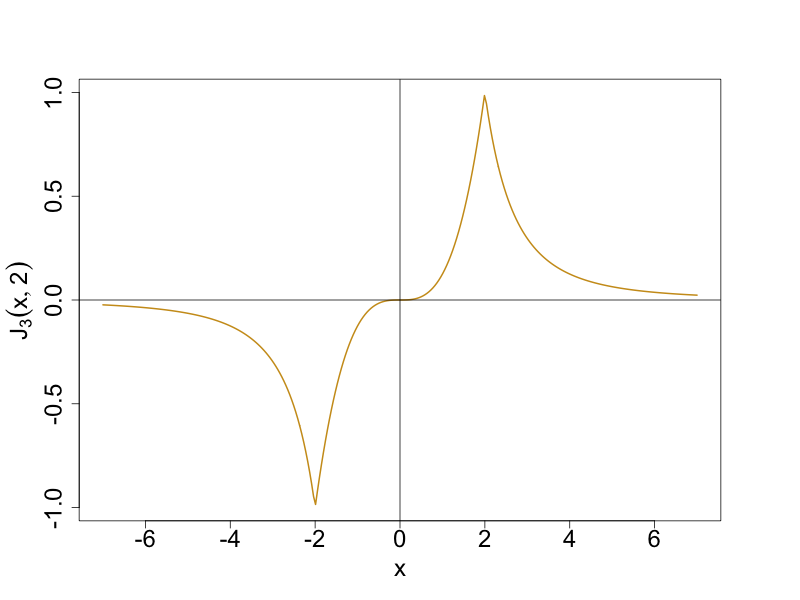}  
    \caption{The similarity function $J_3(x,2)$, when compared to the real-valued Jaccard index,
    is characterized by a sharper peak, therefore implying even more strict similarity quantification.}
    \label{fig:sharper}
    \end{center}
\end{figure}

Figure~\ref{fig:sharper} illustrates this function for $y=2$.

\vspace{0.5cm}

The above development can be generalized to any non-negative integer degree $D$ odd as:
\begin{equation}  \label{eq:JaccD}
  J_D(x, y) = J(x,y)^D
\end{equation}

\begin{figure}[h!]  
\begin{center}
   \includegraphics[width=0.8\linewidth]{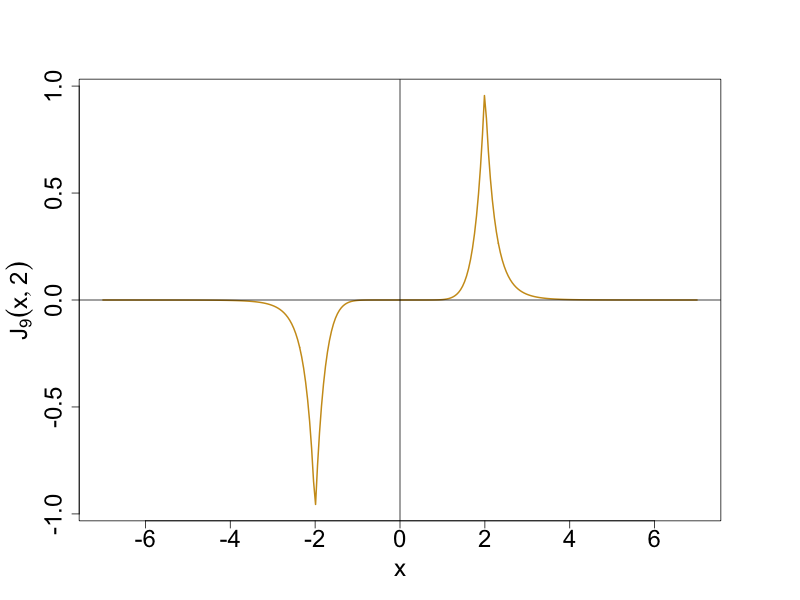}  
    \caption{Similarity quantification through the function $S_{D=9}(x,2)$.}
    \label{fig:even_sharper}
    \end{center}
\end{figure}
\vspace{0.5cm}

An adaptation can be also implemented in case we need to consider $D$ even.  It is also possible
to have less strict similarity comparison by adopting $D<1$, which also requires some adaptation of
Equation~\ref{eq:JaccD}.  

Observe that the similarity function $S_D(x,y)$ tends to the generalized Kronecker delta function
when $D\rightarrow \infty$, $D$ odd, i.e.:
\begin{equation}
  \lim_{D \rightarrow \infty} S_D(x,y)= \tilde{\delta}(x,y)
\end{equation}

However, for simplicity's sake, we will consider only the real-valued Jaccard index
in our subsequent performance analysis, which can be understood as the above
construction when $D=0$.  A more systematic study of higher values of $D$ will be reported 
opportunely.

Given that the Jaccard similarity index has been verified not to be able to take into account
the relative interiority (or overlap, e.g.~\cite{Kavitha}) between the two compared vectors, 
the \emph{coincidence similarity} has been proposed as a respective complementation,
consisting of the product between the real-valued Jaccard and interiority indices:
\begin{equation}
   \mathcal{C}_R(\vec{x},\vec{y}) = \mathcal{J}_R(\vec{x},\vec{y}) \ \mathcal{I}_R(\vec{x},\vec{y})
\end{equation}

Since the coincidence index does not distinguish from the respective real-valued Jaccard index
for the one-dimensional input (i.e.~$x, y, \in \mathcal{R}$), we now assume that the two
values to be compared are vectors in an $N-$dimensional space, i.e.~$\vec{x}, \vec{y} \in \mathcal{R}^N$.

In this case, the real-valued Jaccard index can be expressed as:
\begin{align}
   \mathcal{J}_R(\vec{x},\vec{y}) = \frac{ \sum_{i=1}^N s_{x_i y_i} \min\left\{ s_{x_i}  x_i, s_{y_i} y_i \right\} } {\sum_{i=1}^N  \max\left\{ s_{x_i} x_i, s_{y_i} y_i \right\} }
\end{align}

The interiority index for real valued vectors can be expressed~\cite{CostaSimilarity, CostaJaccard,CostaMset} as:
\begin{align} \label{eq:interiority}
   I(\vec{x},\vec{y}) = \frac{ \sum_{i=1}^N  \min\left\{ s_{x_i} x_i, s_{y_i} y_i \right\} } {\max\left\{ S_{\vec{x}}, S_{\vec{y}} \right\} }
\end{align}

where:
\begin{align}
  S_{\vec{x}} = \sum_{i=1}^N  s_{x_i} x_i \\
  S_{\vec{y}} = \sum_{i=1}^N  s_{y_i} y_i 
\end{align}

It is interesting to observe that both the Jaccard and interiority indices, as with the Euclidean distance, 
are \emph{invariant to permutations} of the indexing $i$ of the input components, in the sense that
any changes in the order of the two input vectors will yield identical results.  The permutation
invariance of similarity or distance indices, which implies that they cannot take into account the sense of
eventual input rotations or reflections, is compatible with the fact that the order of the
features in pattern recognition systems can rarely be specified.

\section{Multiset Similarities in Two-Dimensional Spaces}

Having presented and discussed the properties of the inner product and real-valued Jaccard similarity indices
respectively to comparing two real values $x$ and $y$, we now extend that discussion to two dimensional spaces,
so that now we are interested in comparing the similarity between two real-valued vectors $\vec{x}=[x_1,y_1]^T$ and 
$\vec{y}==[x_2,y_2]^T$, with $\vec{x}, \vec{y} \in \mathcal{R}^2$.

\begin{figure}[h!]  
\begin{center}
   \includegraphics[width=0.8\linewidth]{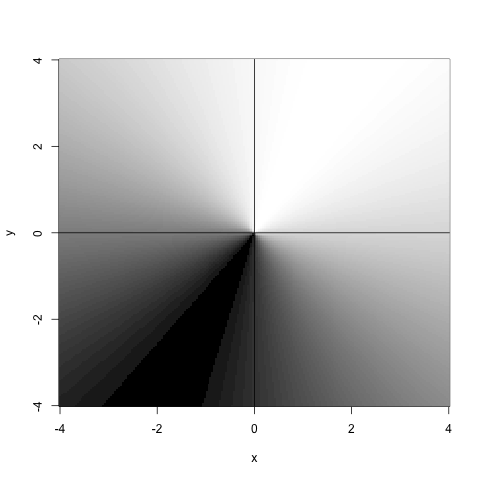}  
    \caption{Cosine similarity values (shown in grayscale from dark to bright) obtained for the cosine similarity
    index between a reference vector $\vec{y} = [1,2]^T$ and vectors $\vec{x} = [x_1,y_1]^T$ with 
    $4 \leq x_1,y_1 \leq 4$.  }
    \label{fig:cosine2D}
    \end{center}
\end{figure}

Figure~\ref{fig:cosine2D} presents the cosine similarity calculated between a reference vector $\vec{y} = [1,2]^T$ and
vectors $\vec{x} = [x_1,y_1]^T$ with $4 \leq x_1,y_1 \leq 4$.

\vspace{0.5cm}

The maximum similarity takes place for the angle sector containing the vector $\vec{y}$,
but any other vector with the same angle will imply identical cosine similarity, therefore illustrating the fact that
the cosine similarity cannot distinguish between any two vectors with the same orientation but distinct magnitudes.
In addition, observe that the gray levels undergo rather little variations for vectors with orientations similar to that
of $\vec{y} = [1,2]^T$.  These issues can have severe impact on the pattern recognition performance of individual neurons
based on the cosine similarity.  Analogous implications are expected for $N-$dimensional input.

\begin{figure}[h!]  
\begin{center}
   \includegraphics[width=0.8\linewidth]{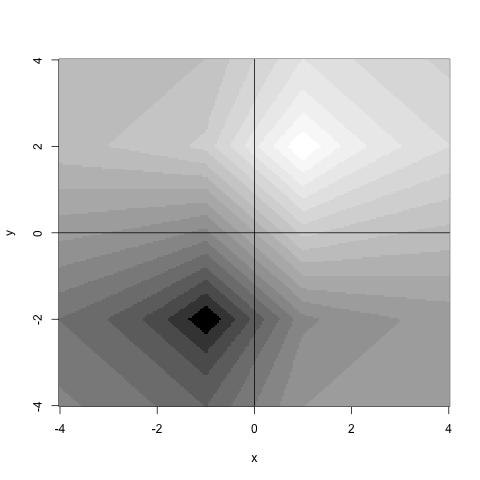}  
    \caption{Jaccard similarity values ($D=1$) obtained when comparing two real-valued vectors $\vec{x}=[x_1,y_1]^T$ and 
              $\vec{y}=[1,2]^T$, with $4 \leq x_1,y_1 \leq 4$.}
    \label{fig:jacc2D}
    \end{center}
\end{figure}

Figure~\ref{fig:jacc2D} depicts the similarity values obtained for the real-valued Jaccard index considering the 
same comparison problem as before.  The surface in this and the subsequent figures in this section are shown
with substantially reduced gray level resolution (comparatively to that in Fig.~\ref{fig:cosine2D}) in order to make the underlying
geometry of the surfaces more discernible in terms of respective level set curves.

\vspace{0.5cm}

Unlike the results obtained for the cosine similarity, now we have a well-defined and delimited peak (bright gray
levels) corresponding to the position $\vec{y}=[1,2]^T$.  As expected, a minimum peak is also observed at $\vec{y}=[-1,-2]^T$.
The enhanced specificity and strictness of the comparison implemented by the real-valued Jaccard index, when compared
to the cosine similarity results in Figure~\ref{fig:cosine2D}, are striking.  As in the one-dimensional case discussed in the
previous section, the real-valued Jaccard similarity index has been able to quantify the input similarity with great accuracy
while preserving a good level of generalization.

\begin{figure}[h!]  
\begin{center}
   \includegraphics[width=0.6\linewidth]{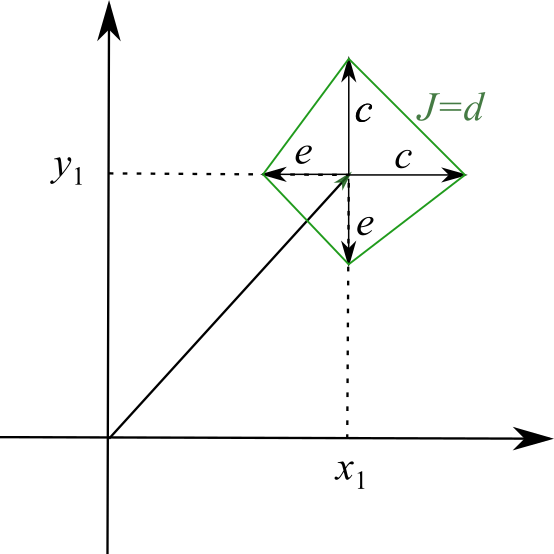}  
    \caption{The basic construction characterizing the equisimilarity region for the
    real-valued Jaccard, defined by the position of vectors  $\vec{v_2} = \left[ x2, y2 \right]^T$
    when compared to a reference vector $\vec{v_1} = \left[ x1, y1 \right]^T$
    and fixed Jaccard similarity value $J=d$.}
    \label{fig:level_set}
    \end{center}
\end{figure}

The Jaccard similarity presents an intrinsic geometry and symmetry worth focusing attention on.  Figure~\ref{fig:level_set}
depicts a diagram of the equisimilarity region defined by making $\mathcal{J}_R(\vec{x},\vec{y})$ equal to a constant
value $d$.  

\vspace{0.5cm}
\vspace{0.5cm}

While keeping $x_1$ fixed, and assuming $y_2 > y_1$, we can write:
\begin{equation}
  d = \mathcal{J}(\vec{v_1},\vec{v_2}) = \frac{y_1 + y_2} {y_1 + y_2 + d} = \frac{y_1 + y_2 - e} {y_1 + y_2}  
\end{equation}

which implies:
\begin{align}
    c  = \frac{ \left( y_1 + y_2 \right) \left(1 - d \right)  }  {d}   \label{eq:c} \\
   e = \left( y_1 + y_2 \right) \left(1 - d \right)   \label{eq:d}
\end{align}

It also follows that:
\begin{equation}
 d =\frac{e}{c}
\end{equation}

meaning that the equisimilarity regions tends to a symmetric diamond when $d \rightarrow 1$.

Now, if we scale both vectors as $\vec{v}_{1,s} = \kappa \vec{v}_{2}$ and $\vec{v}_{2,s} = \kappa \vec{v}_{2}$,
with $\kappa \in \mathcal{R}$, it follows that the dimensions of the respectively scaled region are  $c_s  = \kappa c$
and $e_s = \kappa e$, indicating that the size of the equisimilarity region changes linearly with the scaling of the
vectors being compared.  

\begin{figure}[h!]  
\begin{center}
   \includegraphics[width=0.8\linewidth]{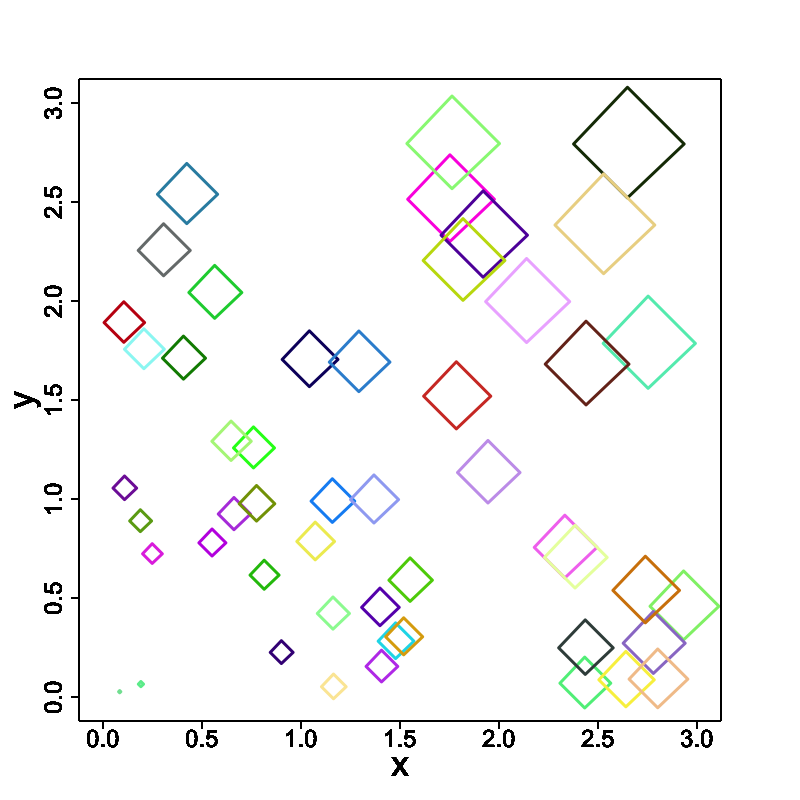}  
    \caption{The real-valued Jaccard similarity index is naturally normalized with respect to the
    scaling of the magnitudes of the compared vectors.  This figure illustrates
    several equisimilarity regions obtained for the real-valued Jaccard similarity
    with $J = d = 0.95$.  Observe the linear scale of the regions size with the position of the reference
    vector.}
    \label{fig:several}
    \end{center}
\end{figure}

In other words, the real-valued Jaccard similarity naturally implements the often sought scaling
invariance, being normalized respectively to the magnitude of the vectors.  This is in contrast to the
Euclidean distance, which has constant equisimilarity region, so that normalization with respect to scale 
requires additional calculations of vector  magnitudes and respective divisions.   Thus, the Jaccard similarity values
are relative to the vectors magnitude.  Figure~\ref{fig:several} illustrates
several equisimilarity regions for the real-values Jaccard similarity.  An analogous property characterizes
the coincidence similarity index, as it derives directly from the Jaccard index.

\vspace{0.5cm}

\begin{figure}[h!]  
\begin{center}
   \includegraphics[width=0.8\linewidth]{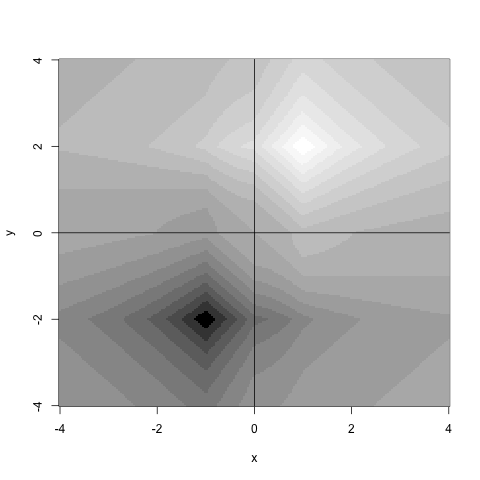}  
    \caption{The coincidence similarity obtained while comparing  two real-valued vectors $\vec{v}_1=[x_1,y_1]^T$ and 
              $\vec{v}_2=[1,2]^T$, with $4 \leq x_1,y_1 \leq 4$.}
    \label{fig:coinc2D}
    \end{center}
\end{figure}

Figure~\ref{fig:coinc2D} presents the coincidence similarity values obtained for the same comparison problem.  An even more
strict comparison can be obtained.  Observe also, in comparison with the surface in Figure~\ref{fig:jacc2D}, the distinct
shape of the level-set contours, which reflect the incorporation of the interiority index into the Jaccard similarity quantification.

\begin{figure}[h!]  
\begin{center}
   \includegraphics[width=0.8\linewidth]{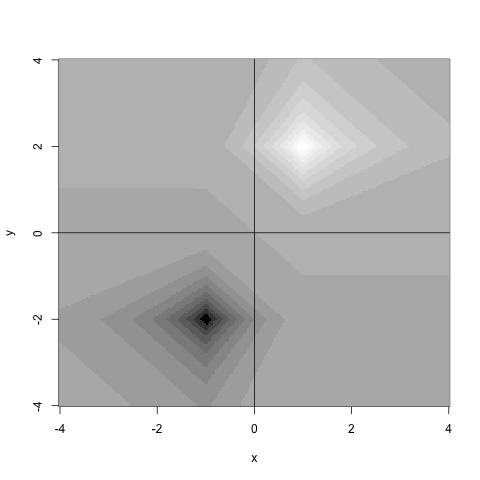}  
    \caption{The Jaccard similarity obtained for $D=9$ while comparing  two real-valued vectors $\vec{v}_1=[x_1,y_1]^T$ and 
              $\vec{v}_2=[1,2]^T$, with $4 \leq x_1,x_2 \leq 4$.}
    \label{fig:sharper2D}
    \end{center}
\end{figure}

Even stricter, sharper comparisons can be obtained by using $D>1$.  Figure~\ref{fig:sharper2D} presents the similarity
surface obtained for the real-valued Jaccard index with $D=9$.  Substantially sharper peaks are obtained at the expense
of reduced generalization capabaility.

\vspace{0.5cm}

As described in~\cite{CostaCCompl}, both the real-valued Jaccard and real-valued coincidence indices can be generalized to
incorporate a parameter $\alpha$, $0 \leq \alpha \leq 1$, controlling the relative contribution of the pairwise features with the same or opposite signs  on the resulting similarity values.   In case $\alpha >0.5$, the contribution of the
pairwise features with the same sign will be enhanced, with the opposite taking place for $\alpha$.  When $\alpha=0.5$,
this index becomes identical to its parameterless version.  The availability of the
parameter $\alpha$ has been verified to enhance the level of details and modularity when of the application of
the coincidence for translating datasets described by respective features into complex networks (e.g.~\cite{CostaCCompl}). 

\begin{figure}[h!]  
\begin{center}
   \includegraphics[width=0.8\linewidth]{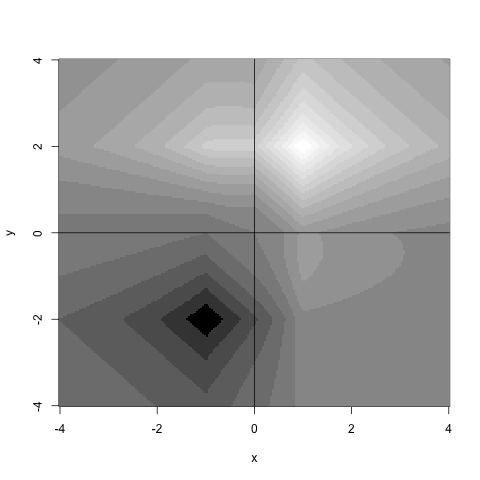}  
    \caption{The coincidence similarity obtained for $D=1$ with $\alpha=0.7$ while comparing  two real-valued vectors $\vec{v}_1=[x_1,y_1]^T$ and 
              $\vec{v}_2=[1,2]^T$, with $4 \leq x_1,y_1 \leq 4$.  The maximum and minimum obtained coincidence values are
              $0.691$ and $-0.296$, respectively, confirming the enhancement of the contribution of pairwise features with the same sign.}
    \label{fig:coinc2D_alpha07}
    \end{center}
\end{figure}

\begin{figure}[h!]  
\begin{center}
   \includegraphics[width=0.8\linewidth]{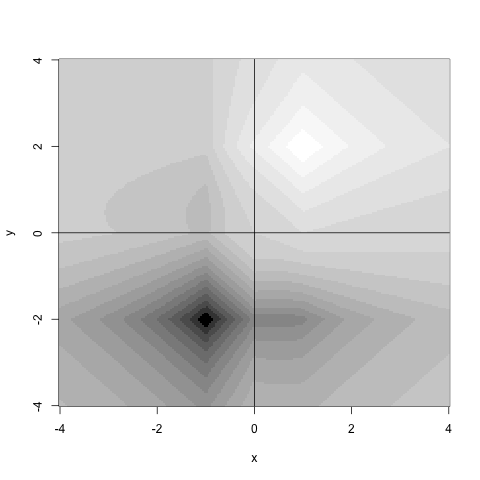}  
    \caption{The coincidence similarity obtained for $D=1$ with $\alpha=0.3$ while comparing  two real-valued vectors $\vec{v}_1=[x_1,y_1]^T$ and 
              $\vec{v}_2=[1,2]^T$, with $4 \leq x_1,y_1 \leq 4$.  The maximum and minimum obtained coincidence values are
              $0.296$ and $-0.691$, respectively, confirming the enhancement of the contribution of pairwise features with the same sign.}
    \label{fig:coinc2D_alpha03}
    \end{center}
\end{figure}

Figures~\ref{fig:coinc2D_alpha07} and Figures~\ref{fig:coinc2D_alpha03} presents the coincidence values obtained for the
same comparison problem as above, but with $\alpha = 0.7$ and $\alpha = 0.3$, respectively.  

\vspace{0.5cm}

\vspace{0.5cm}

The effect of emphasizing the relevance of pairwise features have the same or opposite signs is marked in these figures,
confirming the importance of the additional parameter $\alpha$ in controlling how aligned or anti-aligned pairs of
features are taking into account, which can lead to enhanced comparison details.   Observe also that the adoption of
$\alpha \neq 0.5$ implies in the positive and negative peaks to become asymmetric, with the peak with the lower
magnitude becoming smaller and less sharp.

\section{Sensitivity to Localized Perturbations}

When dealing with the features to be input to a pattern recognition system such as a neuronal cell,
standardization of each of the individual features along its ensemble is often adopted as a means to
normalize the dispersion of each feature so it has null mean and unit standard deviation.  This is
implemented in order to avoid that features taking larger or shifted values predominate over the
other features with smaller magnitude.    In other words, every effort is often made so that no
individual feature dominate the results.  

This same concern extends naturally to the effect of variations of any of the features magnitude on
the overall result.  That is important in cases such in which some of the features are
too noisy or not particularly relevant, which is often the case.

This section addresses the sensitivity of the Jaccard and cosine similarities, as well as the 
normalized Euclidean distance with respect to small perturbations to any of the isolated components
of the two vectors to be compared.

We focus on the situation in whch all $x_i$, $y_i$ are positive, with $y_i >x_i$, $\forall i$, but the
results are similar for the other cases.  In this case, the Jaccard similarity can be simplified as:
\begin{align}
  \mathcal{J}_R(\vec{x},\vec{y}) = \frac{ \sum_{i=1}^N  x_i } {\sum_{i=1}^N  y_i  }
\end{align}

so that the respective variation implied by the small perturbation $\partial x_i$ can be
immediately obtained as:
\begin{align}
   \frac{\partial \mathcal{J}_R(\vec{x},\vec{y})}{\partial x_i} = \frac{ 1} {\sum_{i=1}^N  y_i  }
\end{align}

From the definition of Euclidian distance, we have:
\begin{align}
 E(\vec{x},\vec{y}) = 2  \sqrt{ \sum_{i=1}^N \left( x_i - y_i \right)^2 } 
\end{align}
 
When normalized by the average of the magnitudes of $\vec{x}$ and $\vec{y}$, this distance becomes:
\begin{align}
 E(\vec{x},\vec{y}) = 2  \frac{ \sqrt{ \sum_{i=1}^N \left( x_i - y_i \right)^2 } }   {|| \vec{x} || + || \vec{y} ||  }
\end{align}

Assuming that the variation $\partial x_1$ does not significantly change $|| \vec{x}  ||$, we can
derive  the following approximation:
\begin{align}
 \frac{ \partial E(\vec{x},\vec{y})} {\partial x_i} \approx \frac{x_i - y_i}   {|| \vec{x} || + || \vec{y} ||  }
\end{align}

from which we can infer that small perturbations to any individual component $x_i$ will imply variations of
the normalized Euclidean distance that are not only proportional to the magnitude of $x_i$, but also depend
strongly on $y_i$.

\begin{figure}[h!]  
\begin{center}
   \includegraphics[width=0.9\linewidth]{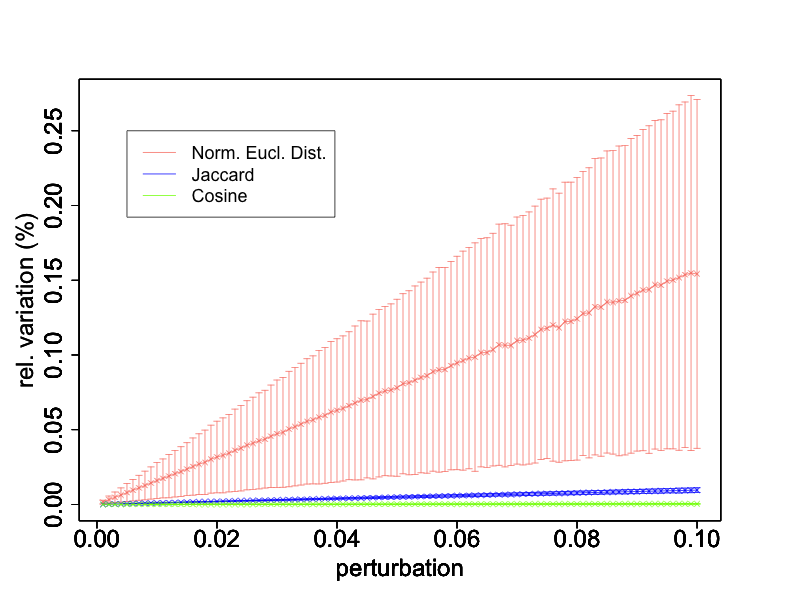}  
    \caption{Relative variations of the Jaccard and cosine similarities, as well as the normalized Euclidean
    distance, to small perturbations of a single component $x_i$ of one of the vectors being compared.
    Values are shown as average $\pm$ standard deviation, in terms of the magnitude of the individual
    perturbations.   The stabilities of the two similarities, regarding both the average and standard deviations of the 
    relative variations, are substantially smaller than that of the normalized Euclidean distance.}
    \label{fig:sensit}
    \end{center}@book{Mirkin,
author = " B. Mirkin",
title = "Mathematical Classification and Clustering",
publisher = "Kluwer Academic Publisher",
address = "Dordrecth",
year = 1996
}

\end{figure}

Figure~\ref{fig:sensit} illustrates the relative variations of the normalized Euclidean distance, as well as
Jaccard and cosine similarities in terms of the magnitude of small perturbations to a single component, 
assuming $N=100$, each of the components of $\vec{x}$ drawn from a normal distribution with average 10
and standard deviation 3, and $\vec{y}$ obtained by adding $\vec{x}$ to $0.1$ multiplied by a vector 
with coordinates drawn from a normal distribution with mean 1 and standard deviation 5.   The values
were obtained from 10000 random experiments.

\vspace{0.5cm}

Though respective to a specific configuration, these results still illustrate that the normalized Euclidean distance
is particularly sensitive to small perturbations on the magnitude of any of the components of the vectors being
compared.  This often represents a substantial shortcoming while comparing vectors and recognizing patterns,
especially with some of the components are particularly noisy or not so much relevant to the pattern analysis.

Given that the sensitivity  of the coincidence similarity can be verified to be comparable to that of the Jaccard 
similarity, and considering that the cosine similarity implements little strict comparisons, we have that the
Jaccard and coincidence similarities tend to have substantial advantages regarding the stability of the obtained
results respectively to perturbations implied by some of the components of the vectors being compared.

\section{Generalized Multiset Neurons}

Traditional implementations of artificial neuronal networks, as integrate-and-fire models as McCulloch and
Pitts and perceptrons (e.g.~\cite{haykin,mcculloch}), often involve the inner product of the input signal with
the respective synaptic weights (bilinear operation) followed by a non-linearity output function.   As such, these neurons
implement linear discrimination.  

\begin{figure}[h!]  
\begin{center}
   \includegraphics[width=0.7\linewidth]{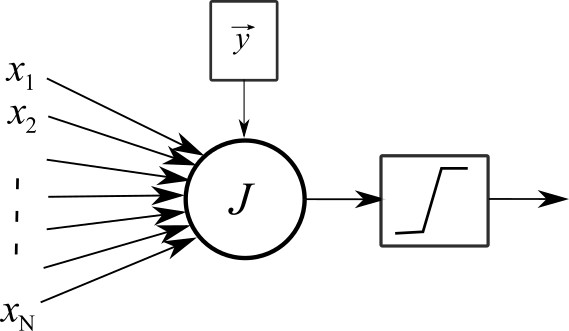}  
    \caption{A generalized multiset neuron with $N$ inputs incorporating a non-linear hardlimiting
    output function.}
    \label{fig:jaccneur}
    \end{center}
\end{figure}

The multiscale neurons described so far in the present work involve a comparison between two vectors, one of
which can be understood as a template, while the other corresponds to the input signal to be compared with the
template.  We have seen in the previous section that these neurons have receptive fields that resemble 
diamonds.   In addition to being able to compare generic templates in any dimension, it is also possible to
use this type of neuron to implement binary decision regions.  This can be accomplished by incorporating a
non-linear output function, as in the integrate-and-fire paradigm that receives the Jaccard value as input, while
the output intensity can be understood as an indication of the certainty of the template recognition.
Figure~\ref{fig:jaccneur}(a) illustrates one neuron of this type (Jaccard similarity) considering $N-$dimensional
input, while some possible $2-$dimensional decision regions illustrated in (b).

\vspace{0.5cm}

\begin{figure}[h!]  
\begin{center}
   \includegraphics[width=0.7\linewidth]{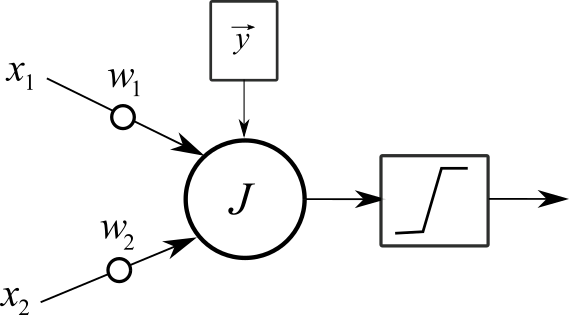}  
    \caption{The \emph{gemini} neuron. The generalization of multiset neurons to incorporate non-linear output function as well as
    synaptic weights associated to each of its inputs.  The comparison part of the neuorn can be based on the real-valued
    Jaccard or coincidence similarities with generic parameter configurations.}
    \label{fig:gemini}
    \end{center}
\end{figure}

Though it may firstly appear that the multiset neurons implement more specific decision regions, they are in fact 
much more generic and versatile than integrate-and-fire neurons.
In order to harness the full potential of multiset neurons, it is necessary to incorporate a small modification in 
the sense that each synaptic input is multiplied by a respective weight, as illustrated in Figure~\ref{fig:gemini}(a),
while some of the possible respective decision regions are exemplified in (b).  These neurons are henceforth
referred to as \emph{generalized multiset neurons}, or \emph{geminis} (GMNs) for short.

\vspace{0.5cm}

The fact that the basic decision region of a multiset neuron is a diamond, while a half-plan is obtained for
integrate-and-fire neurons, allows impressive possibilities for obtained quite elaborate and generic
decision regions by combining the input from just a few geminis.

\section{Single Neuron Comparison}

In this section, we perform a comparison of single neurons defined respectively to the
real-valued Jaccard, interiority, and coincidence indices, as well as to the classic
inner product.  The similarity indices are considered for implementing the synaptic
efficiency and dendritic integration of stimuli up to the implantation cone.  Therefore,
the intrinsic non-linearity of the latter is not considered in this work.  The non-linearity
here is accounted by the multiset-based operations implemented at each synapsis.

This comparison is developed by taking into account several possible effects commonly
found regarding pattern recognition by single neuronal cells, including: (a) relative position
displacements; (b) stimulus size variation (scaling); (c) stimulus intensity variation; (d) noise;
and (e) presence of more than a single pattern in the stimulus.

The reference input stimulus will be a circularly symmetry two-dimensional gaussian 
function centered at the stimulus space, given as:
\begin{eqnarray}  \label{eq:gaussian}
   g(x,y) = e^{-0.5 \left( \frac{d(x,y)}{\sigma} \right)^2}  \\
   \emph{where: }  d(x,y) = \sqrt{ x^2 + y^2}
\end{eqnarray}

Unless otherwise stated, we adopt $\sigma = 100$ in an $200 \times 200$ image support.

\begin{figure}[h!]  
\begin{center}
   \includegraphics[width=0.9\linewidth]{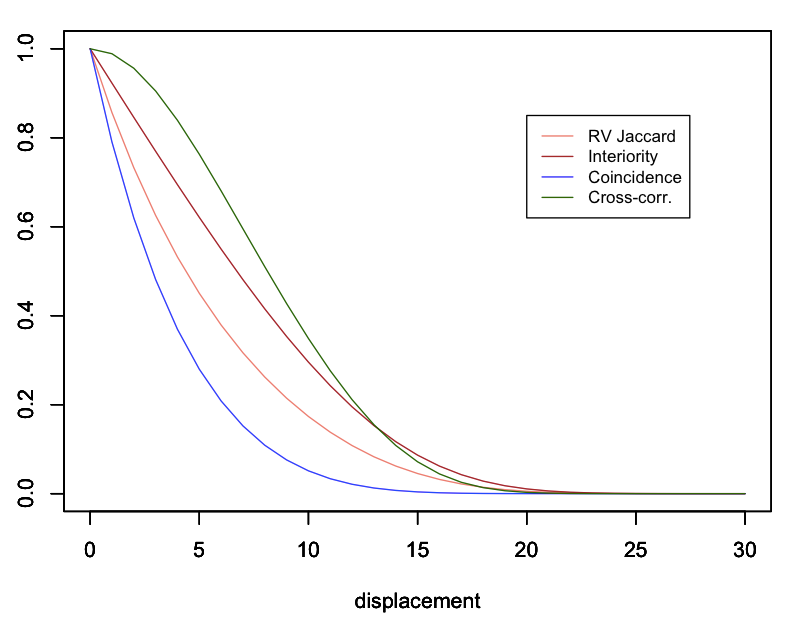}  
    \caption{Similarity values obtained by the four considered methods respectively
    to relative displacements of two identical gaussians.  The coincidence method
    allowed the fastest, and therefore most strict, quantification of the similarity, 
    while the classic cross correlation yielded the most tolerant and least
    discriminative results.}
    \label{fig:displ}
    \end{center}
\end{figure}

Figure~\ref{fig:displ} presents the values of the four considered methods respective
to relative displacements from 0 to 30 discrete steps (pixels).  Full similarity has been
duly identified by all methods regarding null displacement, as could be expected.
However, as soon as one of the patterns shifts, the values of all indices are decreased.
The sharpest decrease is verified for the coincidence approach, which is 
known~\cite{CostaJaccard,CostaSimilarity} to provide a more strict quantification of
pairwise similarity.  

The classic cross correlation presented the slowest decrease
between all methods, except for displacements above 12 pixes, in which case all
the indices values are already very small. This is in agreement with the identification
of product based similarities~\cite{CostaJaccard,CostaSimilarity} to be particularly
tolerant to pairwise differences.  The real-valued Jaccard approach yielded the
second fastest decreasing values.

\vspace{0.5cm}

\begin{figure}[h!]  
\begin{center}
   \includegraphics[width=0.9\linewidth]{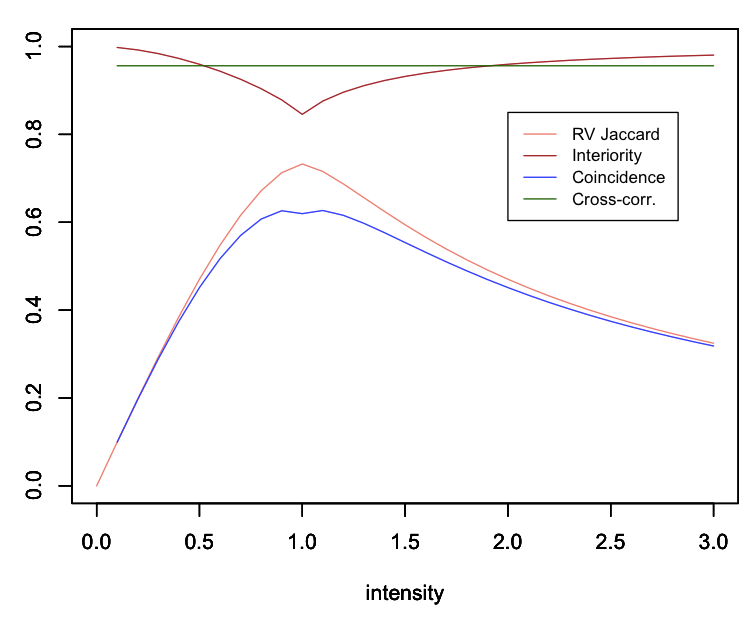}  
    \caption{Quantification of the similarity between two circularly symmetric gaussians
    with the same dispersion, but one of them multiplied by a scaling factor from 0 to 3.
    In order to impose a more challenging demand, one of the gaussians was always
    shifted by 2 pixels along each axis.  The best results are again observed for the
    coincidence method, followed by the real-valued Jaccard and interiority approaches.
    The classic cross correlated revealed to be completely insensitive to the intensity
    changes.}
    \label{fig:intens}
    \end{center}
\end{figure}

Next, we analyze the similarity quantification in terms of varying intensities of one
of the two identical gaussians, though one of them was displaced by 2 pixels
along both axes in order to impose a more challenging similarity quantification.  
The considered intensity changes varied in a range from  0 to 3.  The results are 
depicted in Figure~\ref{fig:intens}.

\vspace{0.5cm}

Particularly interesting results can be discerned from this figure.  Of greatest notice
is the complete insensitivity of the classic cross-correlation method to the intensity
variations.  Though this feature can be helpful in some applications where intensity
variance is desired, it will completely fail in cases where more strict quantifications
of similarity are required to take into account also the relative intensities. 
The best results in this sense have been obtained with respect to the coincidence
methodology, followed by the real-valued Jaccard approach.  The interiority
yielded a counter intuitive result, in the sense that it presented the smallest
value precisely when the two compared patterns have the same intensity.  That is
so because of the $2$ pixels displacements along the two axes.

It is also worth noticing that the two multiset-based methodologies present two
main behaviors.  From intensities ranging from 0 to 1, meaning that one of the
patterns is less intense than the reference, both these methods present an almost
linear increase up to identical intensity.  The maximum similarity value 1 was
not obtained in this case because of the small imposed relative displacement
of 2 pixels along each axes.  From this peak, the similarity values then decrease
progressively as the intensities, which now correspond to magnifications, increase.

\begin{figure}[h!]  
\begin{center}
   \includegraphics[width=0.9\linewidth]{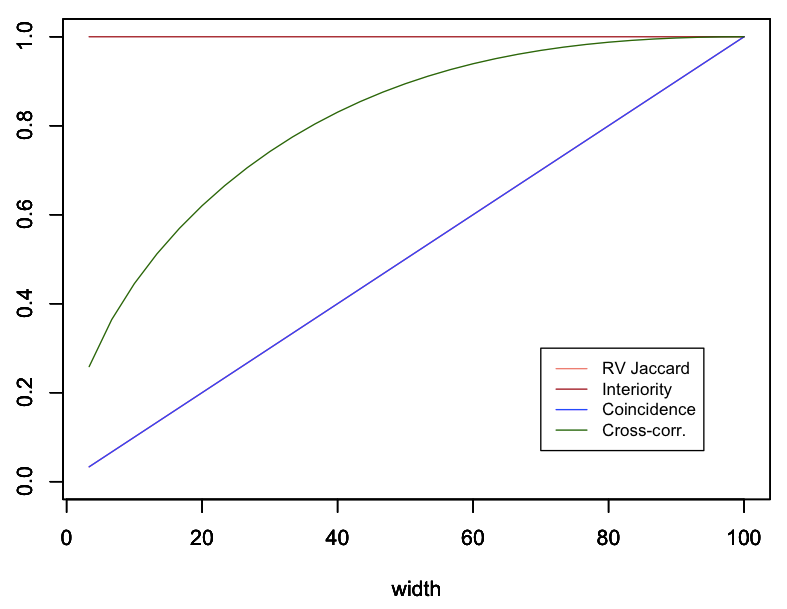}  
    \caption{The similarity between the two patterns in terms of the variation of the
    width, corresponding to the standard deviation, of one of them.  The similarity
    for the real-time Jaccard and coincidence approaches varies linearly from
    0 to 1.  The cross correlation presents a steep initial variation followed by a saturation.
    The kept constant at 1, which is expected given that one of the patterns is always
    interior to the other.}
    \label{fig:width}
    \end{center}
\end{figure}

The results of the study of the effect of the pattern width (or scaling) on the respective matching 
is shown in Figure~\ref{fig:width}.  The width, which corresponded to the standard deviation
of the circularly symmetric gaussian, given in Equation~\ref{eq:gaussian}, varied
from 0 to 100.

\vspace{0.5cm}

Both the real-valued Jaccard and the coincidence match values presented a linear
increase from 0 to 1.  Recall that, in this experiment, both compared patterns correspond to 
circularly symmetric gaussians with $\sigma=100$ pixels.  

The classic cross-correlation presented an initially steep
increasing profile followed by a saturation.  As expected, the interiority index
was kept constant with value 1, reflecting the fact that one of the patterns is
always interior to the other in this particular experiment.   The real-valued Jaccard
and coincidence indices represent a suitable choice in case the similarity is to
reflect the width discrepancy in a linear manner.  The classic cross correlation again
resulted more tolerant to the implemented variation, reaching relatively high values
sooner than the multiset-based methods.

We now proceed to the consideration of additive symmetric uniform noise to one
of the patterns.  More specifically, the following noise levels are added:
\begin{equation}
    X\left[x,y\right] = X\left[x,y\right] +  \frac{i}{N_{ns}} \left[ u(x,y) -0.5 \right]
\end{equation}

with $i = 0, 1, \ldots, N_{ns}$ and where $u(x,y)$ is a scalar uniform random field
taking values in $[0.1]$.  We henceforth adopt $N_{ns} = 20$.  A total of 
20 experiments were performed for each of these levels, the respective
average and standard deviation being then considered as results.

\begin{figure}[h!]  
\begin{center}
   \includegraphics[width=0.9\linewidth]{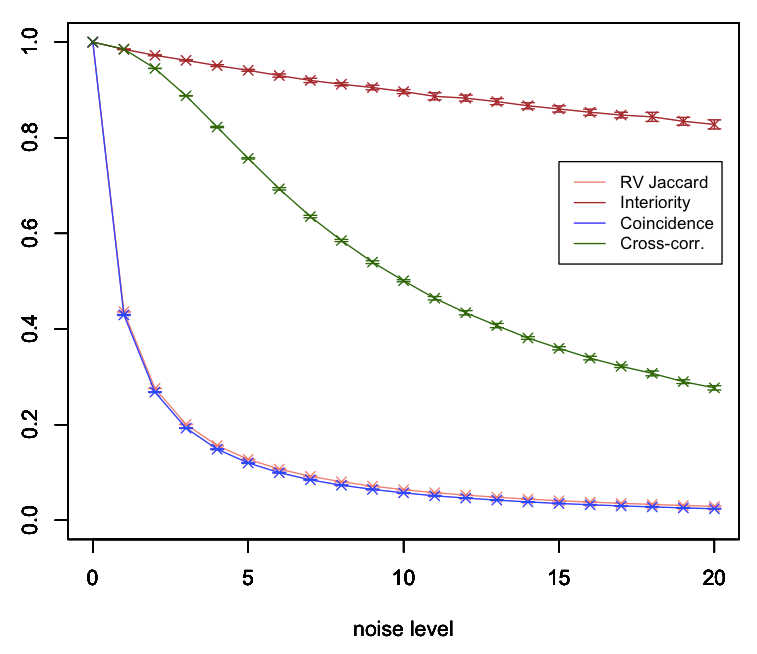}  
    \caption{Similarity values obtained by the four considered methods respectively
    to increasing noise levels.  The interiority approach is the most tolerant, followed
    by the classic cross-correlation, and then the two multiset-based methods.
    Though these curves correspond to respective averages $\pm$ standard deviations,
    the latter are generally very small to be visualized.}
    \label{fig:noise}
    \end{center}
\end{figure}

Figure~\ref{fig:noise} illustrates the similarity values obtained by the four methods
with respect to increasing levels of noise.  Several aspects of interest can be
identified from this figure.  First, we have that the interiority similarity accounts for
the slowest decreasing similarity values.  This can be explained by the fact that
the noisy versions of one of the patterns, despite being jagged, will be mostly
interior to the other.  

The fastest decreasing profiles are those obtained for the
real-valued Jaccard and coincidence methods, which also resulted very similar
one another.  This indicates that these two multiset-based approaches are the
most sensitive to the pattern modifications induced by the increasing levels of noise.
The classic cross-correlation yielded an intermediate result between the interiority
and multiset-based methods, again reflecting its increase tolerance to perturbations.

\vspace{0.5cm}

\begin{figure}[h!]  
\begin{center}
   \includegraphics[width=0.9\linewidth]{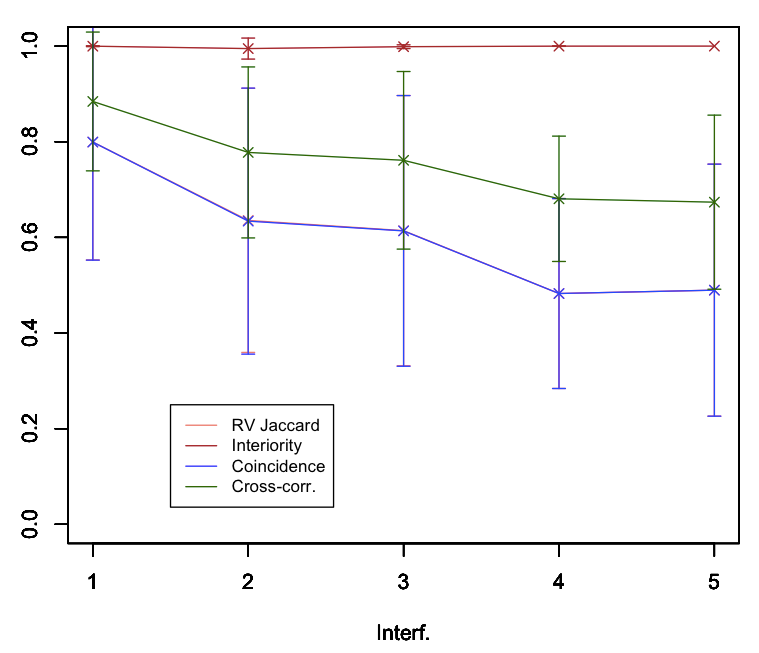}  
    \caption{Similarity values obtained in presence of added interference corresponding to
    signed addition of from 1 to 5 gaussian patterns at uniformly random positions in the
    image. The curves correspond to the average $\pm$ standard deviations.  Identical
    results have been obtained for the real-valued Jaccard and coincidence based
    similarities.}
    \label{fig:interf}
    \end{center}
\end{figure}

The last considered type of perturbation concerns the signed addition of whole gaussian
patterns into one of the images.  From 1 to 5 such patterns have been added into one
of the images at uniformly random positions.  The patterns can be added while being
multiplied by $+1$ or $-1$, chosen in uniformly random manner.
The results are shown in Figure~\ref{fig:interf}.

\vspace{0.5cm}

While the interiority and classic cross-correlation presented total tolerance to the added
interference, a moderate discrimination can be observed in the case of the real-valued
Jaccard and coincidence results, which present total overlap in the figure.

All in all, the several analysis reported in this section further substantiated the tendency
of the multiset-based approaches to provide a more accurate and discriminative 
quantification of the stimulus recognition than the interiority and cross-correlation based
methods.  Significant differences in the specificity of the response have been observed
in several cases, especially varying intensities, widths, and noise.   Given their markedly
more strict and discriminative characteristics, the real-valued and Jaccard, and even more
so the coincidence approach, therefore correspond to the best choice, among the
considered possibilities, for implementing more strict pattern recognition with high levels of
accuracy.

There is an important issue to be further discussed here, and it regards the interplay
between discriminative and tolerant (or invariant) performance.  One first important point concerns
the fact that these seem to be opposite properties, in the sense that a neuron that is too 
tolerant will provide no specific response, and vice versa.  Another critical issue
concerns the fact that,  taken independently, neither of these two properties are necessarily
good or bad.  As in an engineering problem, the best solution will be that which best
suits the specific requirements.

However, in the context of effective recognition of several types of patterns in typical applications,
in presence of all the considered perturbations, perhaps the most proper solution is a 
balanced combination of discriminative and tolerant abilities.  Actually, there is a 
formal solution to this duality between specificity and generality that is not so often
realized.  It concerns the fact that it is indeed possible to achieve both characteristics
in a synergistic manner, not as a kind of trade-off or balance.    This solution consists
of having sets of neurons, each of which highly discriminative and specific, whose
combined operation provides for the requested levels of tolerance and generalizations. 

Thus, while each instance of the presented pattern will be accurate and specifically
identified by successive individual cells, at the overall group
level  substantial tolerance will be achieved for several instances and perturbations
of the presented stimuli.  Nevertheless, this ideal architecture can only be achieved
at expense of substantial informational resources, be then biological or artificial.
These flexible \emph{and} highly discriminative ensembles, which constitute the
ideal solution for many circumstances, are henceforth denominated 
\emph{synergistic neuronal systems}.

The immediate consequence of the above considerations is that it becomes critical
to have the means for implementing strict, discriminative pattern recognition at the
smallest informational and energetic expenses.  From this perspective, the multised-based
similarity identification constitute a particularly interesting resource given their
conceptual and informationally simple operation, allied to their substantially more
strict and discriminative operation as verified in this work with respect to several 
perturbations and in~\cite{CostaComparing} with respect to coexisting patterns.

Interestingly, it has been proposed recently
that the multiset operations can be implemented in extremely efficient manner in analog
electronics, using only a few operational amplifiers and analog switches~\cite{CostaElectronic},
which makes the mutiset-based approaches, and in particular the coincidence index,
components of choice for the development of real-time pattern recognition systems.

It remains an issue of great interest to contemplate how befitted for implementation
in biological hardware the multiset operations ultimately are.

\section{Strictness Effect}

In this section we study the effect of having more strict coincidence comparison, controlled
by the parameter $D$, on the similarity values obtained with respect to the considered
input perturbations.

\begin{figure}[h!]  
\begin{center}
   \includegraphics[width=0.9\linewidth]{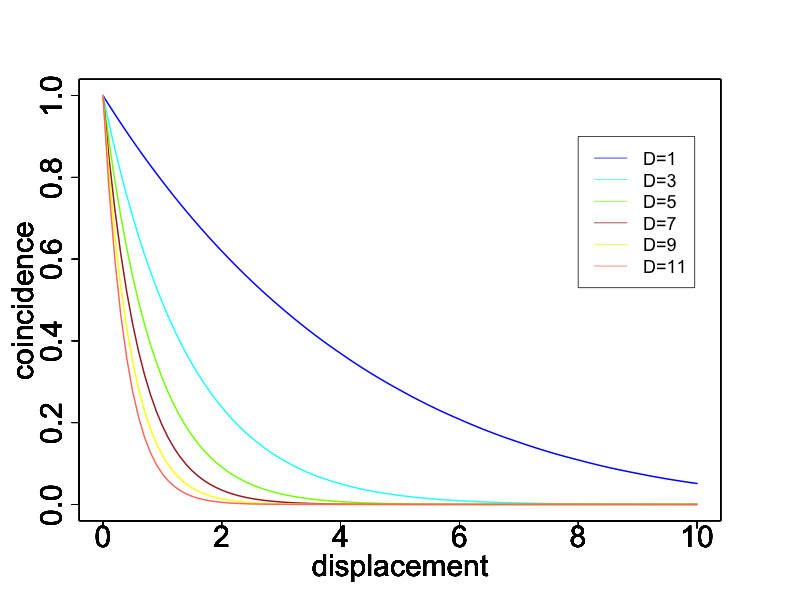}  
    \caption{Similarity values obtained by the coincidence similarity for
    $D = 1, 3, 5, 7, 9, 11$ respectively  to relative displacements of two identical gaussians.  }
    \label{fig:displD}
    \end{center}
\end{figure}

Figure~\ref{fig:displD} illustrates the coincidence values obtained for $D=1, 3, 5, 7, 9, 11$ respectively
to relative displacement between the reference and input gaussians.  It can be readily verified that
the obtained coincidence values decrease steadily with $D$, indicating that progressively more
strict comparisons imply in reducing the obtained coincidence values.

\vspace{0.5cm}

\begin{figure}[h!]  
\begin{center}
   \includegraphics[width=0.9\linewidth]{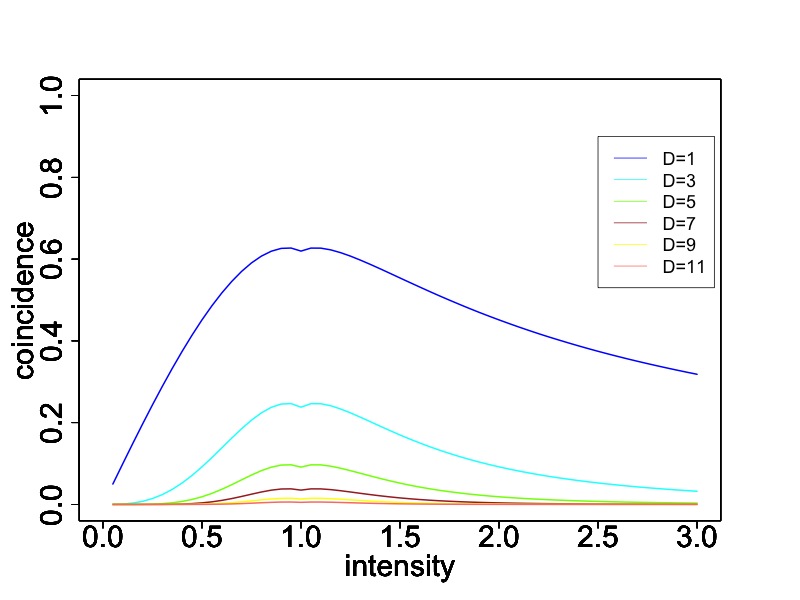}  
    \caption{Quantification of the similarity between two circularly symmetric gaussians
    with the same dispersion, but one of them multiplied by a scaling factor from 0 to 3.
    The similarity values correspond to the coincidence index for $D = 1, 3, 5, 7, 9, 11$.}
    \label{fig:intensD}
    \end{center}
\end{figure}

\begin{figure}[h!]  
\begin{center}
   \includegraphics[width=0.9\linewidth]{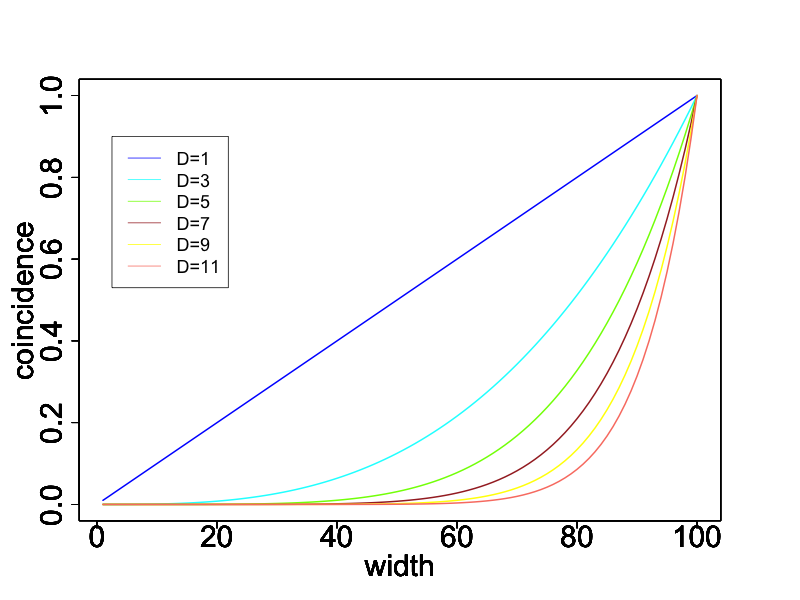}  
    \caption{The coincidence similarity between the two patterns in terms of the variation of the
    width, corresponding to the standard deviation, of one of them.}
    \label{fig:widthsD}
    \end{center}
\end{figure}

The effect of increasing $D$ on the coincidence values obtained with respect to modification of
the intensity of the input is shown in Figure~\ref{fig:intensD}.  Again, the results indicate smaller
indications of the presence of the reference pattern as the coincidence becomes more strict.
Similar effect can be verified respectively to variation of the input width (Fig.~\ref{fig:widthsD}), 
noise and interference.

\vspace{0.5cm}

\vspace{0.5cm}

\section{Application Example: Image Segmentation}

Among the several methods typically involved in image analysis, the segmentation of the
objects of interest constitutes what is possibly the most difficult and challenging task (e.g.~\cite{shapebook,Gonzalez}).
Basically, given an image containing several objects, as well as possibly a background,
the task of \emph{segmentation} consists in identifying the regions in the image that
correspond to the objects of interest.  Observe that image segmentation therefore
corresponds to a \emph{pattern recognition} problem in which each image pixel is
to be classified as belonging or not to the objects of interest.

\begin{figure}[h!]  
\begin{center}
   \includegraphics[width=0.7\linewidth]{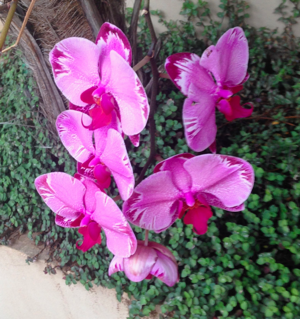}  
    \caption{The image considered for illustrating the suggested multiset neuron based segmentation method.
    This RGB image has size $300 \times 319$ pixels, containing flowers against a varied background that
    includes small leaves with varying hues, intensity, contract and focus.  The identification of the
    green leaves constitutes a particularly challenging problem in image processing and analysis. }
    \label{fig:flower}
    \end{center}
\end{figure}

In the present section we illustrate the impressive potential of multiset neurons based on
the real-valued Jaccard and coincidence similarity indices, for performing supervised image segmentation.
For generality's sake, we will consider the RGB (color) image with size $300 \times 319$ pixels
shown in Figure~\ref{fig:flower}.
The objects of interest will consist of the small leaves in the background of the image.
Observe that these leaves have intense variation of hues, intensities, contrast and even focus,
which contribute to making this problem a particular challenge.

\vspace{0.5cm}

The method proposed here, which is as simple as it is powerful, consists of taking a few samples
of typical pixels belonging to the objects of interest, and then taking the R, G, and B values.
A multiset neuron with hard limit output (see Fig.~\ref{fig:jaccneur})  will be assigned to recognize each of these $N_s$ samples.
The template vector of each of these neurons corresponds to the RGB values of the sampled
pixel as well as of its $w$ neighbors (e.g.~within a square of size $2w+1$).  Having thus
trained the system, the image segmentation proper consists of obtaining the real-valued Jaccard
or coincidence values by comparing the template with each of the image pixels for each of the
multiset neurons, and the resulting similarity value is then thresholded by $T$.  
A logical or is then performed between the obtained outputs of the $N_s$ neurons,
and the original image pixel is understood to belong to the objects of interest whenever at least
one of the neurons yields a true value.

\begin{figure}[h!]  
\begin{center}
   \includegraphics[width=0.7\linewidth]{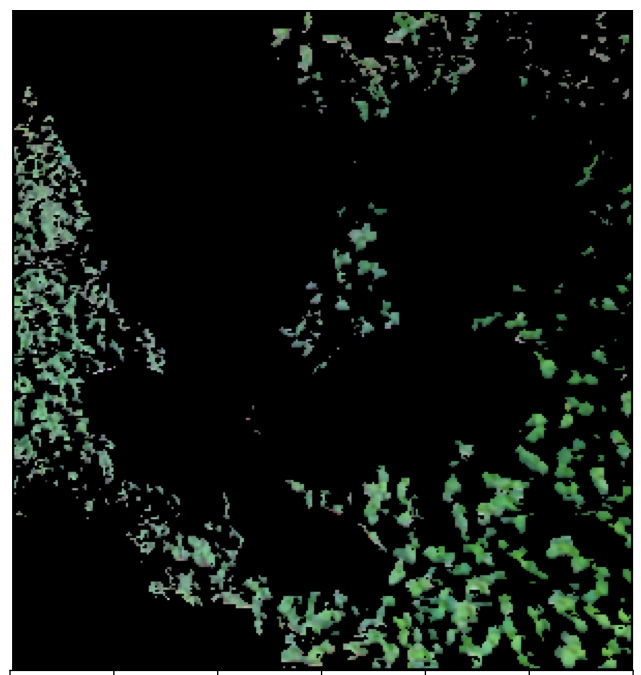}  \\ (a) \\ \vspace{0.2cm}
   \includegraphics[width=0.7\linewidth]{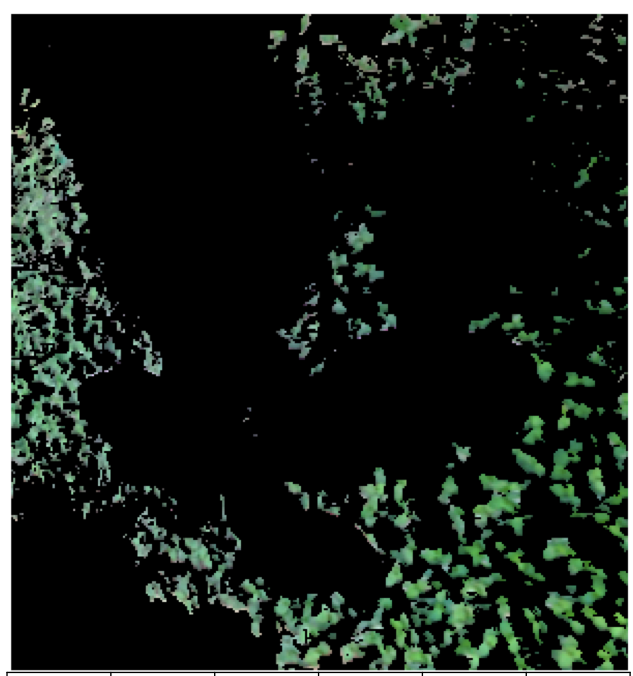}  \\ (b) \\
    \caption{Segmentation of the small leaves in Fig.~\ref{fig:flower} obtained by using 
    multiset neurons based on real-valued Jaccard (a) and coincidence (b) similarity. 
    Only 5 samples (pixels) belonging to the objects of interest were used to train the
    system, which adopted $T=0.8$ and $T=0.75$ respectively to the Jaccard and
    coincidence approaches.  The time required for training is just a few seconds, and 
    the whole identification of the regions takes less than one minute in a standard
    personal computer.}
    \label{fig:segm}
    \end{center}
\end{figure}

Figure~\ref{fig:segm} illustrates the results obtained by taking just $N_s=5$ pixels samples
taken mostly at the lower right portion of the image, respectively to the real-valued Jaccard multiset 
neurons (a) with $T=0.80$, and coincidence multiset neurons (b) with $T=0.75$.  For comparison's 
sake, a smaller threshold has been adopted in
the latter case, since the coincidence is more strict than the Jaccard similarity.   The processing,
which involves just the parameter $T$, took less than one minute in a standard personal computer.

\vspace{0.5cm}

Though similar, the results obtained for the Jaccard and coincidence similarity present some differences
that are consequence of the more strict quantification of the similarity between the pixel properties
which is characteristic of the coincidence approach.  Both results can be considered to be particularly
satisfactory, involving minimal computational resources.

\section{Concluding Remarks}

The present work has developed a study of the application of the multiset-derived
similarity operations, especially the real-valued Jaccard and coincidence indices, to 
artificial neurons.   More specifically, these indices are considered  for substituting 
the inner product performed between the image stimulus and respective matrix of 
synaptic weights. 

After presenting an overview of the related multiset concepts and developments that
led to the real-valued Jaccard and coincidence index, including new results regarding
higher order respective versions, we proceeded to a systematic comparison of artificial
neurons performing pattern recognition in presence of several types of perturbations.
More specifically, the pattern to be recognized is stored in the synaptic weights, while
the similarity comparison is performed by using the several considered indices.

The results largely confirm the enhanced potential of the coincidence index, followed by the
real-valued Jaccard index, for performing strict similarity quantification.  This makes
these types of artificial neurons primary choices for implementations and applications
involving strict patter recognition.  The duality between specificity and generality in 
this type of task has also been discussed, and it has been argued that the ideal solution
is to have large ensembles of highly specific and strict neurons, each of which adapted
for taking into account specific geometric transformations so as to allow respective
invariance.  

Now, a particularly interesting issue arises regarding the fact that, given the substantial
advantages of neurons based on the coincidence or real-valued Jaccard indices, why
would they have not been adopted in biological neuronal networks aimed at effective
pattern recognition?    Why would the otherwise much less efficient inner product be instead
implemented by the dendritic integration of the synaptic input?

There are at least two possible answers to this important question. First, we have that the
biological hardware would be intrinsically unsuitable for implementing the multiset-related
operations.  Interestingly, recent developments have shown that these operations
can be very effectively implemented in analog electronics~\cite{CostaElectronic}, but
this does not necessarily extend to biology, though much of the neuronal operation
is a correlate of electric and even electronic counterparts.  If it happens that biology
is intrinsically unsuitable for performing multiset operations, these alternatives remain
still valid for implementations in other types of hardware.

The second possible answer is that the biological neuronal cells actually implement
multiset-related functions.  Indeed, consider the profile of the operation $x \sqcap y$
shown in Figure~\ref{fig:common}, which is the basis for all effective indices developed
and applied in the present work.  This function, which resembles a sigmoid, 
could be applied not at the implantation cone, but at each of the synapses.  Indeed,
the observed saturation could correspond to the saturation of the synaptic activation
and/or of the local polarization of the interior of the cell.  The sum corresponding to the
numerator of Equation~\ref{eq:Jaccard} would then correspond to the combination
of the diffusive charge effect at the implantation cone.   

As for the denominator of that
same equation, it is possible that other intracellular mechanisms are activated by
the synaptic activity that effectively contribute to the inhibition of the action potential.
These inhibitory effects could be similarity integrated at the implantation cone,
accounting for the denominator in Equation~\ref{eq:Jaccard}.   There are other
possible mechanisms that could account for the implementation of multiset-like
neuronal operations.  For instance, the denominator of Equation~\ref{eq:Jaccard}
could correspond to inhibitory effects received from other cells associated to the
same receptive field that would therefore counterbalance the net depolarization
of the excitatory cell implementing the numerator integration.

Though these are
currently hypothetical, further consideration and experimental developments can
help verifying these possibilities. 

The concepts, methods, and results reported in the present work have several
potential implications in a wide range of areas --- including neuroscience, pattern
recognition and deep learning --- therefore paving the way to a large
number of further developments.  Some examples include further studies of the
possible relationships with biological cells, the consideration of other types of
stimuli, as well as the evaluation of the here introduced higher order versions of the
real-valued Jaccard and coincidence indices.

\vspace{0.7cm}
\emph{Acknowledgments.}

Luciano da F. Costa
thanks CNPq (grant no.~307085/2018-0) and FAPESP (grant 15/22308-2).  
\vspace{1cm}

\bibliography{mybib}

\begin{thebibliography}{10}

\bibitem{haykin}
S.~Haykin.
\newblock {\em Neural Networks And Learning Machines}.
\newblock McGraw-Hill Education, 9th edition, 2013.

\bibitem{mcculloch}
Warren Mcculloch and Walter Pitts.
\newblock A logical calculus of ideas immanent in nervous activity.
\newblock {\em Bulletin of Mathematical Biophysics}, 5:127--147, 1943.

\bibitem{HubelWiesel}
D.~H. Hubel and T.~N. Wiesel.
\newblock {\em Brain and Visual Perception: The Story of a 25-Year
  Collaboration}.
\newblock Oxford University Press, Oxford, 2004.

\bibitem{hubel}
D.~Hubel and T.~Wiesel.
\newblock Receptive fields, binocular interaction, and functional architecture
  in the cat's visual cortex.
\newblock {\em Journal of Physiology}, 160:106--154, 1962.

\bibitem{turner}
M.~H. Turner, G.~W. Schwartx, and F.~Rieke.
\newblock Receptive field center-surround interactions mediate
  context-dependent spatial contrast encoding in the retina.
\newblock {\em eLife}, 7:eLife 2018;7:e38841, 2018.

\bibitem{brigham:1988}
E.~O. Brigham.
\newblock {\em Fast Fourier Transform and its Applications}.
\newblock Pearson, 1988.

\bibitem{RaoHwang}
K.~R. Rao, D.~N. Kim, and J.~J. Hwang.
\newblock Integer fast fourier transform.
\newblock In {\em Fast Fourier Transform - Algorithms and Applications. Signals
  and Communication Technology}, pages 111--126. Springer, Dorcrecht, 2010.

\bibitem{oppenheim:2009}
A.~V. Oppenheim and R.~Schafer.
\newblock {\em Discrete-Time Signal Processing}.
\newblock Pearson, 2009.

\bibitem{Parr:2013}
C.~Phillips, J.~Parr, and E.~Riskin.
\newblock {\em Signals, Systems and Transforms}.
\newblock Pearson, 2013.

\bibitem{Cajal}
S.~Ramon y~Cajal.
\newblock {\em Recollections of My Life}.
\newblock The MIT Press, Cambridge, Mass., 1996.

\bibitem{Friedman}
R.~Friedman.
\newblock Measurements of neuronal morphological variation across the rat
  neocortex.
\newblock {\em Neuroscience Letters}, 734, 2020.

\bibitem{Grueber}
W.~B. Grueber, C.-H. Yang, B.~Ye, and Y.-N. Jan.
\newblock The development of neuronal morphology in insects.
\newblock {\em Current Biology}, 730--738, 2005.

\bibitem{CostaSimilarity}
L.~da~F. Costa.
\newblock On similarity.
\newblock
  \url{https://www.researchgate.net/publication/355792673_On_Similarity}, 2021.
\newblock [Online; accessed 21-Aug-2021].

\bibitem{CostaJaccard}
L.~da~F. Costa.
\newblock Further generalizations of the {J}accard index.
\newblock
  \url{https://www.researchgate.net/publication/355381945_Further_Generalizations_of_the_Jaccard_Index},
  2021.
\newblock [Online; accessed 21-Aug-2021].

\bibitem{CostaCCompl}
L.~da~F.~Costa.
\newblock Coincidence complex networks.
\newblock \url{https://iopscience.iop.org/article/10.1088/2632-072X/ac54c3},
  2022.
\newblock J. Phys.: Compl, 3 015012.

\bibitem{Jaccard1}
P.~Jaccard.
\newblock Distribution de la flore alpine dans le bassin des dranses et dans
  quelques r\'egions voisines.
\newblock {\em Bulletin de la {S}oci\'et\'e vaudoise des {S}ciences
  {N}aturelles}, 37:241--272, 1901.

\bibitem{jac:wiki}
Wikipedia.
\newblock Jaccard index.
\newblock \url{https://en.wikipedia.org/wiki/Jaccard_index}. [Online; accessed
  10-Oct-2021].

\bibitem{CostaMset}
L.~da~F. Costa.
\newblock Multisets.
\newblock \url{https://www.researchgate.net/publication/355437006_Multisets},
  2021.
\newblock [Online; accessed 21-Aug-2021].

\bibitem{Hein}
J.~Hein.
\newblock {\em Discrete Mathematics}.
\newblock Jones \& Bartlett Pub., 2003.

\bibitem{Knuth}
D.~E. Knuth.
\newblock {\em The Art of Computing}.
\newblock Addison Wesley, 1998.

\bibitem{Blizard}
W.~D. Blizard.
\newblock Multiset theory.
\newblock {\em Notre Dame Journal of Formal Logic}, 30:36---66, 1989.

\bibitem{Blizard2}
W.~D. Blizard.
\newblock The development of multiset theory.
\newblock {\em Modern Logic}, 4:319--352, 1991.

\bibitem{Thangavelu}
P.~M. Mahalakshmi and P.~Thangavelu.
\newblock Properties of multisets.
\newblock {\em International Journal of Innovative Technology and Exploring
  Engineering}, 8:1--4, 2019.

\bibitem{Singh}
D.~Singh, M.~Ibrahim, T.~Yohana, and J.~N. Singh.
\newblock Complementation in multiset theory.
\newblock {\em International Mathematical Forum}, 38:1877--1884, 2011.

\bibitem{CostaComparing}
L.~da~F. Costa.
\newblock Comparing cross correlation-based similarities.
\newblock
  \url{https://www.researchgate.net/publication/355546016_Comparing_Cross_Correlation-Based_Similarities},
  2021.
\newblock [Online; accessed 21-Oct-2021].

\bibitem{kumar:2009}
K.~S.~S. Kumar.
\newblock {\em Electric Circuits and Networks}.
\newblock Pearson Education India, 2009.

\bibitem{mirkin}
B.~Mirkin.
\newblock {\em Mathematical Classification and Clustering}.
\newblock Kluwer Academic Publisher, Dordrecth, 1996.

\bibitem{Akbas1}
C.~E. Akbas, A.~Bozkurt, M.~T. Arslan, H.~Aslanoglu, and A.~E. Cetin.
\newblock L1 norm based multiplication-free cosine similiarity measures for big
  data analysis.
\newblock In {\em IEEE Computational Intelligence for Multimedia Understanding
  (IWCIM)}, France, Nov. 2014.

\bibitem{Akbas2}
C.~E. Akbas, A.~Bozkurt, A.~E. Cetin, R.~Cetin-Atalay, and A.~Uner.
\newblock Multiplication-free neural networks.
\newblock In {\em Signal Processing and Communications Applications Conference
  (SIU)}, Malatya, Turkey, May. 2015.

\bibitem{CostaGenMops}
L.~da~F.~Costa.
\newblock Generalized multiset operations.
\newblock \url{https://www.researchgate.net/profile/Luciano-Da-F-Costa}, 2021.
\newblock [Online; accessed 10-Nov-2021].

\bibitem{CostaElectronic}
L.~da~F. Costa.
\newblock Multiset signal processing and electronics.
\newblock
  \url{https://www.researchgate.net/publication/355954430_Multiset_Signal_Processing_and_Electronics},
  2021.
\newblock [Online; accessed 21-Nov-2021].

\bibitem{Kavitha}
M.~K. Vijaymeena and K.~Kavitha.
\newblock A survey on similarity measures in text mining.
\newblock {\em Machine Learning and Applications}, 3(1):19--28, 2016.

\bibitem{shapebook}
L.~da~F.~Costa.
\newblock {\em Shape Classification and Analysis: Theory and Practice}.
\newblock CRC Press, Boca Raton, 2nd edition, 2009.

\bibitem{Gonzalez}
R.~C. Gonzalez and R.~E.~Woods and.
\newblock {\em Digital Image Processing}.
\newblock Pearson, New York, 2018.

\end{thebibliography}
\bibliographystyle{unsrt}

\end{document}